\documentclass[10pt,twocolumn,letterpaper]{article}

\usepackage{iccv}
\usepackage{times}
\usepackage{epsfig}
\usepackage{graphicx}
\usepackage{amsmath}
\usepackage{amssymb}

\usepackage{array} %
\usepackage{graphicx}
\usepackage{amsmath}
\usepackage{amssymb}
\usepackage{relsize}
\usepackage{booktabs}
\usepackage{bm}
\usepackage{blindtext}
\usepackage{balance}
\usepackage{tikz}
\usetikzlibrary{spy,arrows}

\newcommand{\norm}[1]{ \left\lVert#1\right\rVert }
\newcommand{\density}{\sigma}
\newcommand{\px}{\mathbf{x}}
\newcommand{\R}{\mathbb{R}}
\newcommand{\sdf}{d}
\newcommand{\radiance}{L}
\newcommand{\img}{I}
\newcommand{\accradiance}{\mathcal{L}}
\newcommand{\pdf}{w}
\newcommand{\dx}[1]{\;\mathrm{d}#1}
\newcommand{\pn}{\mathbf{n}}
\newcommand{\pc}{\mathbf{c}}
\newcommand{\pp}{p}
\newcommand{\vo}{\mathbf{v}}
\newcommand{\vi}{{\bm\omega}}
\newcommand{\brdf}{f_\text{r}}
\newcommand{\brdfd}{\rho}

\newcommand{\rough}{\alpha_\text{r}}
\newcommand{\spec}{\alpha_\text{s}}
\DeclareMathOperator{\PSF}{\text{PSF}}
\newcommand{\volsdfgeoparam}{\varphi}
\newcommand{\volsdfradparam}{\psi}
\newcommand{\sdfparam}{\theta}
\newcommand{\brdfparam}{\gamma}
\newcommand{\diffuseparam}{\brdfparam_1}
\newcommand{\specparam}{\brdfparam_2}
\newcommand{\e}{\text{e}}

\newcommand{\dogA}{\textit{dog1}}
\newcommand{\dogB}{\textit{dog2}}
\newcommand{\girlA}{\textit{girl1}}
\newcommand{\girlB}{\textit{girl2}}
\newcommand{\pony}{\textit{pony}}
\newcommand{\dragon}{\textit{dragon}}

\newcommand{\bird}{\textit{bird}}
\newcommand{\squirrel}{\textit{squirrel}}

\renewcommand{\S}{\mathbb{S}}
\renewcommand{\(}{\left(}
\renewcommand{\)}{\right)}

\newcommand{\supervol}{SupeRVol}
\newcommand{\nosr}{noSR}

\newcommand{\paperTitle}{\supervol{}: Super-Resolution Shape and Reflectance Estimation\\
in Inverse Volume Rendering}

\newcommand{\paperIDICCV}{1834}

\author{Mohammed Brahimi$^1$ \hspace{2mm}
Bjoern Haefner$^1$ \hspace{2mm}
Tarun Yenamandra$^1$ \hspace{2mm}
Bastian Goldluecke$^2$ \hspace{2mm}
Daniel Cremers$^1$\\
$^1$ Technical University of Munich, $^2$ University of Konstanz\\
{\tt\small$\lbrace$mohammed.brahimi, bjoern.haefner, tarun.yenamandra, cremers$\rbrace$@tum.de}\\ {\tt\small bastian.goldluecke@uni-konstanz.de}
}

\usepackage[pagebackref=true,breaklinks=true,letterpaper=true,colorlinks,bookmarks=false]{hyperref}

\iccvfinalcopy %

\def\iccvPaperID{\paperIDICCV} %

\begin{document}

\title{\paperTitle}

\maketitle
\vspace*{-1cm}
\begin{abstract}
   We propose an end-to-end inverse rendering pipeline called SupeRVol that allows us to recover 3D shape and material parameters from a set of color images in a super-resolution manner.
   To this end, we represent both the bidirectional reflectance distribution function's (BRDF) parameters and the signed distance function (SDF) by multi-layer perceptrons (MLPs).
   In order to obtain both the surface shape and its reflectance properties, we revert to a differentiable volume renderer with a physically based illumination model that allows us to decouple reflectance and lighting.
   This physical model takes into account the effect of the camera's point spread function thereby enabling a reconstruction of shape and material in a super-resolution quality.
   Experimental validation confirms that SupeRVol achieves state of the art performance in terms of inverse rendering quality.
   It generates reconstructions that are sharper than the individual input images, making this method ideally suited for 3D modeling from low-resolution imagery.
\end{abstract}

\section{Introduction}
\label{sec:Intro}
\begin{figure}[t]
  \centering
  \small
  \newcommand{\mywidthc}{0.02\textwidth}
  \newcommand{\mywidthx}{0.143\textwidth}
  \newcolumntype{C}{ >{\centering\arraybackslash} m{\mywidthc} }
  \newcolumntype{X}{ >{\centering\arraybackslash} m{\mywidthx} }
  \newcommand{\tabelt}[1]{\hfil\hbox to 0pt{\hss #1 \hss}\hfil}
  \setlength\tabcolsep{1pt} %
  \begin{tabular}{CXXX}
    \rotatebox{90}{Shape}&
    \includegraphics[width=\mywidthx]{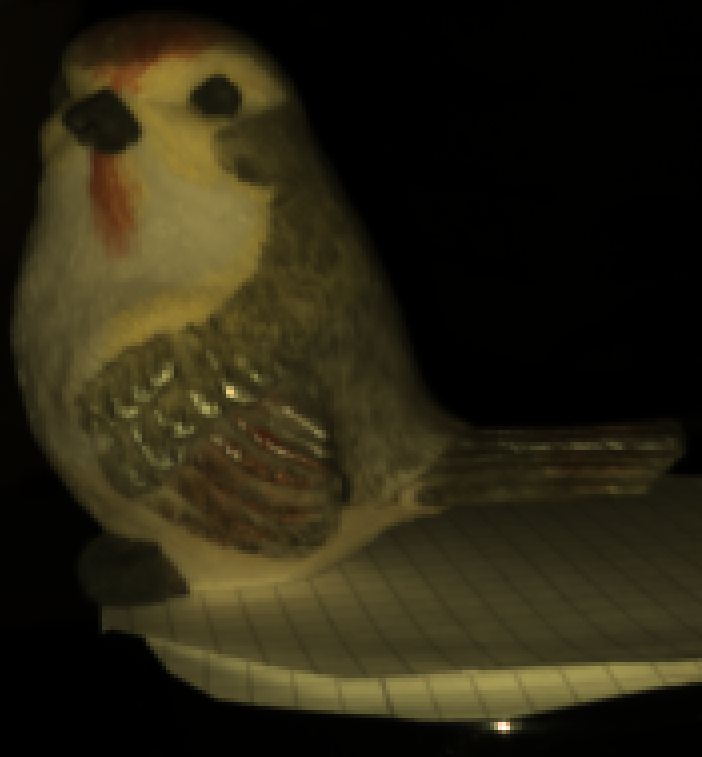}&
    \includegraphics[width=\mywidthx]{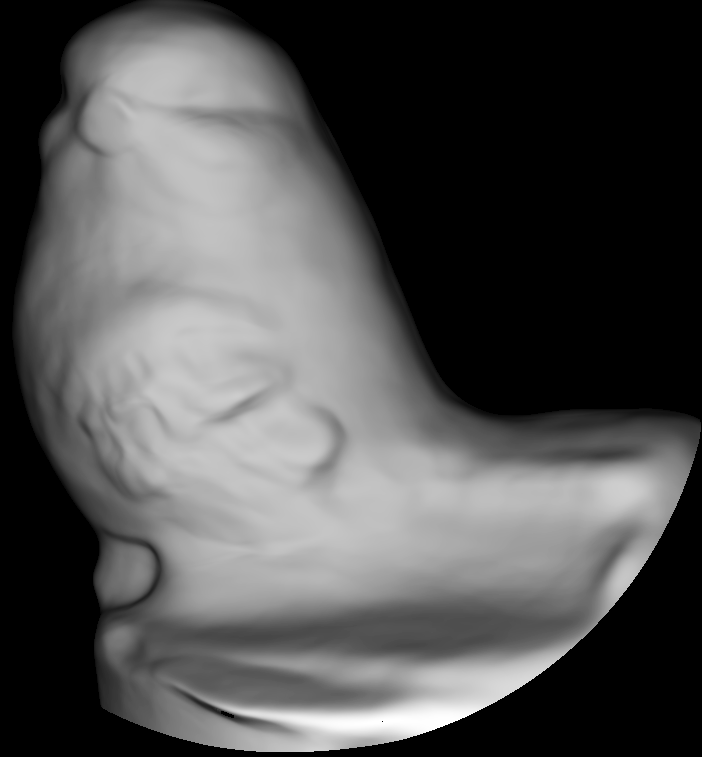}&
        \includegraphics[width=\mywidthx]{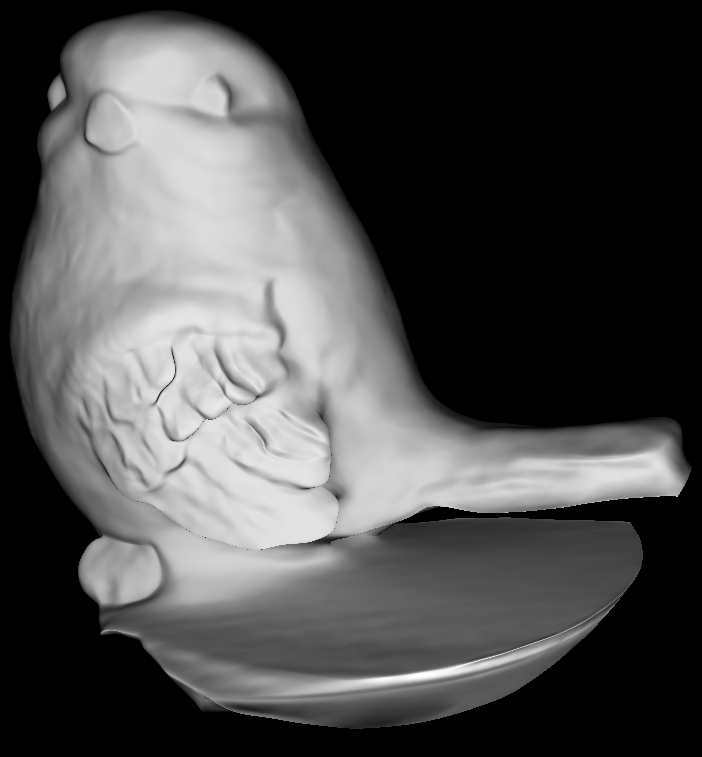}\\
    \rotatebox{90}{Full reconstruction}&
    \begin{tikzpicture}[spy using outlines={rectangle,connect spies}]
      \node[anchor=south west,inner sep=0]  at (0,0) {\includegraphics[width=\mywidthx]{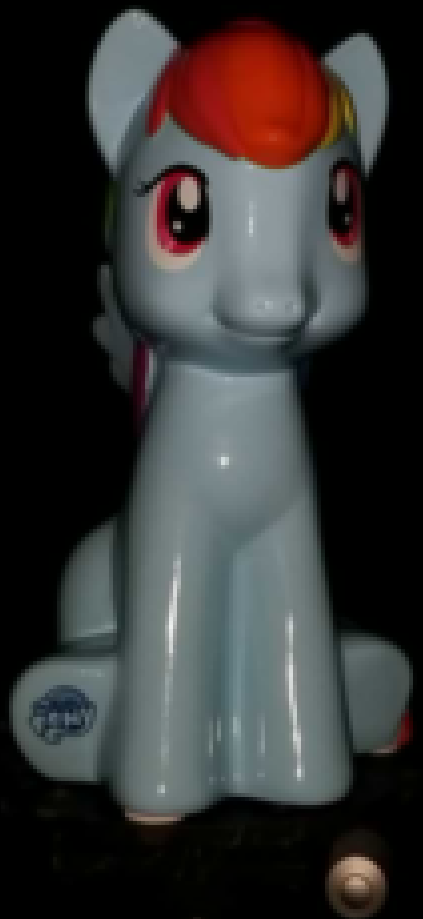}};
      \spy[color=green,width=1.35cm,height=1.2cm, magnification=3] on (0.37,1.215) in node [right] at (1.15,0.6);
    \end{tikzpicture}&
    \begin{tikzpicture}[spy using outlines={rectangle,connect spies}]
      \node[anchor=south west,inner sep=0]  at (0,0) {\includegraphics[width=\mywidthx]{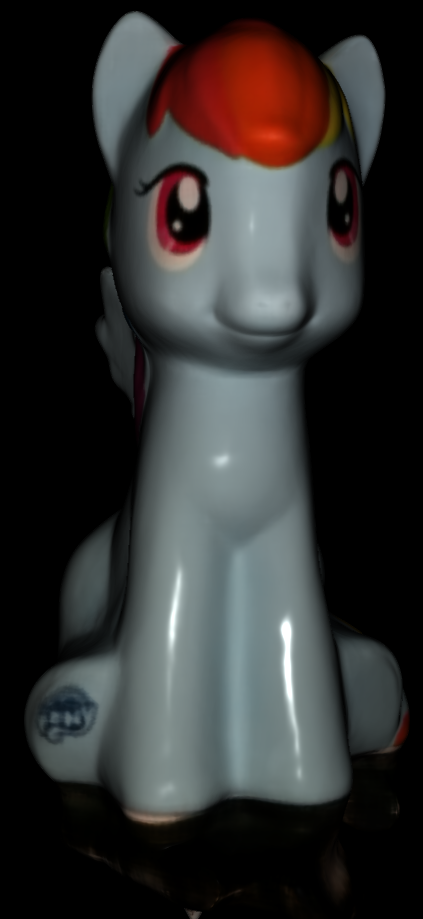}};
      \spy[color=green,width=1.35cm,height=1.2cm, magnification=3] on (0.37,1.215) in node [right] at (1.15,0.6);
    \end{tikzpicture}&
    \begin{tikzpicture}[spy using outlines={rectangle,connect spies}]
      \node[anchor=south west,inner sep=0]  at (0,0) {\includegraphics[width=\mywidthx]{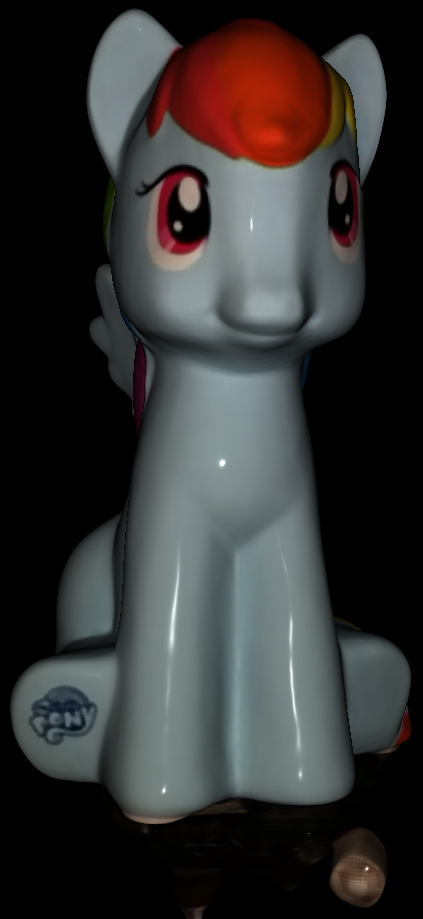}};
      \spy[color=green,width=1.35cm,height=1.2cm, magnification=3] on (0.37,1.215) in node [right] at (1.15,0.6);
    \end{tikzpicture}\\
    & LR input & IRON~\cite{yariv2021volume} & SupeRVol
  \end{tabular}
  \caption{Given a set of low-resolution input images (left column), SupeRVol recovers the geometry (top) and the material properties (full reconstruction at the bottom).   The combination of a realistic physical camera model with inverse volumetric rendering gives rise to  reconstructions that are more crisp than competing ones and even sharper than the input images.}
\label{fig:teaser}
\end{figure}

The reconstruction of 3D shape and appearance is among the classical challenges in computer vision.
While we have witnessed significant progress on this task with suitably designed neural representations, the resulting reconstructions of shape and appearance are typically limited by the resolution of the input images where high-quality models invariably require high-resolution input images.
At the same time, the concept of super-resolution modeling has been studied intensively in classical variational optimization approaches.
The aim of this work is to bring both of these developments together and introduce super resolution modeling into neural differentiable volume rendering approaches in order to allow high-resolution reconstructions of 3D shape and reflectance even from low-resolution input images -- see Figure~\ref{fig:teaser} for high quality geometry and super-resolution image reconstruction on two real world datasets.

More specifically, we consider a setting where we capture images of an opaque non-metallic object from various viewpoints. The so called \textit{photometric} images are illuminated only from a white point light source colocated with the camera. We want to separately recover geometry and different components of an isotropic BRDF.
Note that due to the nature of the setup, we can only reconstruct the slice of the BRDF for which illumination direction is equal to viewing direction.
However, we will see that we generalize well to novel relighting scenarios.
The images are assumed to be calibrated, i.e. camera ex- and intrinsics are given, e.g. from COLMAP~\cite{schonberger2016pixelwise}; %
this is similar to~\cite{mildenhall2021nerf,zhang2021physg,zhang2022iron}.
\\
In detail, our contributions are as follows: 
\begin{itemize}
    \item Given a set of photometric images, we put forward an end-to-end inverse rendering approach for jointly estimating high quality shape of arbitrary topology and its corresponding reflectance properties.
    \item Within the image formation model, we explicitly parameterize the degradation process induced by the camera's sensor via modeling its point spread function, allowing us to estimate super-resolved shape and material properties.%
    \item In numerous experiments, we validate that the proposed method gives rise to state-of-the-art reconstructions of shape and reflectance.
    In particular, the reconstructed objects are significantly more detailed and sharper than the individual input images. 
\end{itemize}

\section{Related Work}
\label{sec:Related_work}
In the following we recall neural inverse rendering and view synthesis as well as super-resolution approaches for 3D reconstruction.

\subsection{Neural Inverse Rendering and View Synthesis}

Neural approaches for inverse rendering~\cite{kaya2021uncalibrated, kellnhofer2021neural,mueller2022instant,sengupta2019neural,taniai2018neural,tewari2022advances,tewari2020state,zhang2022modeling} and novel view synthesis~\cite{gao2021dynamic,hedman2021baking,li2021neural,liu2021neural,mildenhall2021nerf,peng2021neural,riegler2021stable,tancik2022block,wizadwongsa2021nex} have gained a lot of attention over the recent years.
Their expressivity within a lightweight architecture such as a multilayer perceptron (MLP) form a great basis for these complex tasks~\cite{liu2020neural,martin2021nerf,mildenhall2021nerf, pumarola2021d,wang2022nerf,yu2021pixelnerf,zhang2020nerf++}.
However, those approaches can only recover geometry of reduced quality due to their underlying volume rendering
based on the scene's density.
In contrast, surface rendering approaches~\cite{niemeyer2020differentiable,yariv2020multiview} perform better when estimating geometry, but require mask supervision and can still get stuck at unsatisfactory local minima with severe geometry artefacts. This limits their usage to relatively simple objects.
To get the best out of both worlds, \cite{oechsle2021unisurf, wang2021neus, yariv2021volume} express the underlying shape implicitly via occupancy fields~\cite{oechsle2021unisurf} or SDFs~\cite{wang2021neus, yariv2021volume}.
\cite{yariv2021volume} even provide a sampling procedure of the volume integral to theoretically upper bound the opacity error, which is a missing desired property in~\cite{oechsle2021unisurf, wang2021neus}.
While~\cite{oechsle2021unisurf, wang2021neus, yariv2021volume} allow to reliably reconstruct accurate geometry of complex scenes without mask supervision, these approaches lack the important ability to estimate other scene properties such as reflectance.

Oftentimes it is desired to recover the object's BRDF properties along with the geometry, as this allows for relighting the scene under novel illumination.
While some works focus on relighting based on static, unknown illumination of the scene~\cite{boss2021nerd,srinivasan2021nerv,zhang2021nerfactor, zhang2021physg}, others enforce a change of illumination~\cite{bi2020deep,luan2021unified,nam2018practical,zhang2022iron} by resorting to %
photometric images. %
In particular, this can lead to a well-posed optimization problem of the underlying scene properties~\cite{brahimi2020springer}.

Density based volume rendering approaches such as~\cite{bi2020deep,srinivasan2021nerv,zhang2021nerfactor,boss2021nerd} also estimate material parameters, but they inherit the same limitations as~\cite{mildenhall2021nerf}: 
inaccurate geometry and the fact that the object's material parameters are not properly defined on the surface, but everywhere in the volume.
This severely limits the editing capability, and a volume rendering step has to be used during inference, making it unusable in common graphics pipelines.
\cite{luan2021unified,nam2018practical} are mesh-based classical inverse rendering approaches and require masks and a good initialization of the geometry, causing their convergence to be fragile, as mesh-based optimization is inherently non-differentiable at depth discontinuities and difficult to handle if topological changes or self-intersections arise.
More recently, \cite{zhang2022iron} proposed an inverse rendering approach which eliminates the disadvantages of mesh-based approaches to some extent due to the use of a neural SDF representation and an edge-aware surface renderer.
However, the weak convergence properties of surface rendering made them resort to a two-step approach, where the first step consists of a volume rendering step~\cite{wang2021neus} which is used to initialize the second step based on surface rendering.

Compared to prior works using photometric images, our approach is based solely on volume rendering using SDFs, causing reliable geometry and reflectance reconstruction 
without resorting to other rendering techniques such as surface rendering, thus avoiding a multi-step pipeline. Additionally, thanks to our novel problem formulation our method is still applicable for standard graphics pipelines, as a mesh and each surface point's reflectance property can be easily recovered.
Next, we discuss the state-of-the-art in super-resolution for 3D reconstruction and how we leverage that to further improve our methodology.

\subsection{Super-Resolution for 3D Reconstruction}

The problem of super-resolution (SR) has been extensively studied in the past~\cite{anwar2020deep,nasrollahi2014super,park2003super,tian2011survey,van2006image,wang2020deep,yang2019deep,yue2016image}. Different problem statements of SR lead to different approaches, e.g. the case of single image SR~\cite{glasner2009super,huang2015single,niu2020single}, video SR~\cite{chan2021basicvsr,chan2022basicvsr++,isobe2022look}, or depth SR~\cite{riegler2016atgv,haefner2019photometric,sang2020inferring,voynov2019perceptual,yang2007spatial}. Given that we are interested in a 3D reconstruction of the scene from a set of photometric images, we do not perform SR in 2D image space, but in 3D scene space~\cite{goldlucke2014super,maier2015super,wang2022nerf}.
\cite{wang2022nerf} are able to synthesize images of higher resolution than the individual input images by resorting to supersampling, i.e. a low-resolution pixel is sampled at each super-resolution pixel's center, allowing for a denser sampled radiance field.
\cite{goldlucke2014super,maier2015super} are classical approaches optimizing SR geometry and textures.
While~\cite{maier2015super} integrate low-resolution depth and color from and RGB-D sensor into SR keyframes and fuse these keyframes into a texture map,~\cite{goldlucke2014super} describe their SR process using a convolution with a Gaussian kernel.
This is a well motivated image formation model of a camera sensor element and straightforward to carry out as they work in a discrete pixel grid.
However, this is not easily applicable in a continuous case, i.e. when using neural networks to implicitly express shape and reflectance. 
While~\cite{goldlucke2014super,maier2015super,wang2022nerf} share the benefit of increasing the input resolution of the individual input images to result in a sharper, high detailed output, they all lack the possibility of representing the scene's intrinsic properties, i.e. shape and material.
Either the full scene is represented in a neural network~\cite{wang2022nerf} or the estimated textures consist of lighting cues baked in to the reflectance properties, making faithful relighting impossible~\cite{goldlucke2014super,maier2015super}.

Contrary to the existing SR works mentioned here, we mathematically formulate the camera's image formation process in the continuous case leading to a principled sampling heuristic applicable for neural approaches which allows us to invert the camera's image formation model resulting in reconstructions beyond the input in terms of resolution and quality.
Additionally, our model can recover super-resolved geometry and BRDF parameters, free from baked in lighting cues, allowing faithful photorealistic reconstructions with full control over the scene's properties.
\section{Preliminaries: VolSDF}

\label{sec:Preliminaries}
VolSDF~\cite{yariv2021volume} leverages volume rendering similarly to NeRF~\cite{mildenhall2021nerf}, however aims at overcoming certain limitations of NeRF by decoupling geometric representation and appearance.
To this end, VolSDF models the scene geometry within a volume $\Omega\subset\R^3$ by means of a density~$\density:\Omega\to\R_{\geq0}$, which is related to its signed distance function (SDF)
$\sdf:\Omega\to\R$ by the transformation
\begin{equation}
\label{eq:preliminaries:density}
\density(\px) = \alpha\Psi_\beta(-\sdf(\px)).
\end{equation}
Here, $\Psi_\beta$ is the Cumulative Distribution Function of the Laplace distribution with zero mean and scale $\beta$, and both~$\alpha, \beta>0$ are learnable parameters.
\begin{figure}[t]
  \centering
  \small
  \newcommand{\mywidthx}{0.235\textwidth}
  \newcolumntype{X}{ >{\centering\arraybackslash} m{\mywidthx} }
  \newcommand{\tabelt}[1]{\hfil\hbox to 0pt{\hss #1 \hss}\hfil}
  \setlength\tabcolsep{1pt} %
  \begin{tabular}{XX}
    \includegraphics[width=\mywidthx]{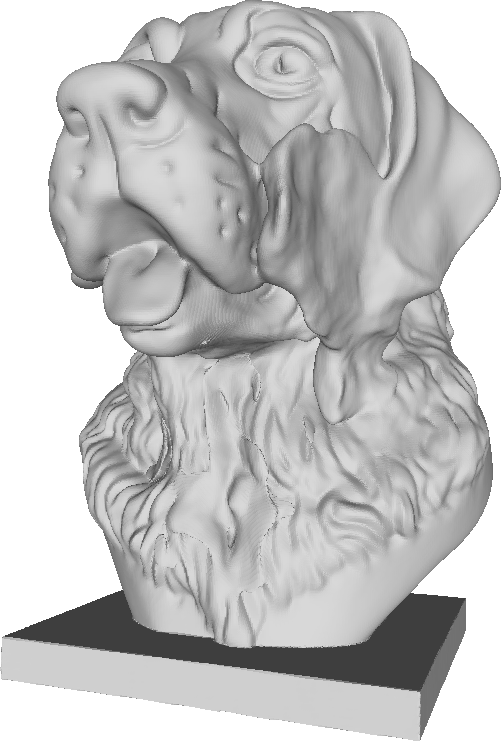}&
    \includegraphics[width=\mywidthx]{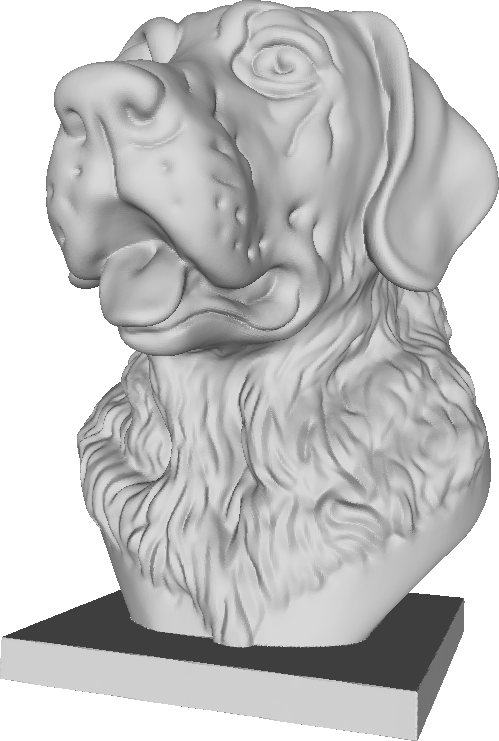}\\
    Surface Rendering & Volume Rendering
  \end{tabular}
  \caption{Geometric reconstruction using our approach based on two different rendering strategies. Surface based rendering~\cite{yariv2020multiview} gets easily stuck in bad local minima with less detail (hair on neck) and undesirable artifacts (cheek). Our approach based on volume rendering~\cite{yariv2021volume} does not suffer from these issues, resulting in highly detailed reconstructions.}
\label{fig:volvssurf}
\end{figure}

Given this parameterization of the density in terms of the underlying SDF, we can set up the volume rendering equation to obtain radiance~$\accradiance_\pp$ at a pixel~$\pp\in\R^2$ within the image of a camera located at~$\pc\in\R^3$. Let~$\vo\in\S^2$ be the viewing direction from~$\pc$ through~$\pp$, then 
\begin{equation}
  \accradiance_\pp = \int_{0}^\infty \pdf(t) \radiance(\px(t),\pn(t),\vo)\dx{t}
  \label{eq:volume_integral},
\end{equation}
where we integrate along the ray~$\px(t) = \pc + t \vo, t \in \R$.
The weights~$\pdf(t)$ form a probability distribution along the ray~\cite{yariv2021volume},
and are given by %
\begin{equation}
  \pdf(t)=\density(\px(t))\,\exp\left( -\int_0^t\density(\px(s))\dx{s} \right ).
\label{eq:weights}
\end{equation}
Finally, $\radiance:\Omega\times\S^2\times\S^2\to\R^3$ is the radiance field, which depends on location,
normal and viewing direction.
Since positive values of the SDF,~$d$ inside the surface are assumed, the normal vector is obtained as %
$\pn=\nabla \sdf/\norm{\nabla \sdf}$. %
In practice, the integral~\eqref{eq:volume_integral} is being approximated using the well-known quadrature rule at a discrete set of samples
$t_1 < t_2 < \dots < t_{m}$
for each pixel,
\begin{equation}
  \accradiance_p \approx \sum^{m-1}_{i=1} (t_{i+1}-t_i)\pdf(t_i)L(\px(t_i),\pn(t_i),\vo).
  \label{eq:discrete_volume_integral}
\end{equation}
Note, that the integral in~\eqref{eq:weights} to compute~$\pdf(t_i)$ is accumulated in a similar way while iteratively computing the sum~\eqref{eq:discrete_volume_integral}.
VolSDF represents the scene using two separate MLPs, one is used to describe the SDF, $\sdf_{\volsdfgeoparam}$ \textit{and} a global geometry feature map $z_\volsdfgeoparam:\Omega\to\R^{256}$, while a second MLP is used to describe the radiance $\radiance_{\volsdfradparam}$, both with their corresponding learnable network parameters $\volsdfgeoparam,\volsdfradparam$.
Additionally to its stable convergence compared to surface rendering approaches~\cite{niemeyer2020differentiable,yariv2020multiview}, cp. Figure~\ref{fig:volvssurf}, VolSDF satisfies a theoretical guarantee to upper bound the opacity error compared to similar approaches~\cite{oechsle2021unisurf, wang2021neus}.
Specifically, after convergence the rendered image showing the geometry of the SDF using a volume renderer is almost indistinguishable from the same rendered image based on a surface renderer using e.g. a sphere tracing algorithm~\cite{hart1996sphere}.
This observation was shown in~\cite{oechsle2021unisurf} and has the consequence that the object can be rendered using standard surface renderer frameworks~\cite{li2018differentiable,nimier2019mitsuba,zhang2020path}.
Additionally, the appearance is then defined \textit{on} the object's surface, despite being learned as a volumetric quantity.
Nevertheless, VolSDF is unable to recover the reflectance properties, since both the material and lighting are baked into the radiance network.
Hence, the reconstructed 3D model can only be rendered with the same material under the same static illumination.
In the next section, we show how we extend VolSDF to enable joint estimation of shape and material, which allows material editing and view synthesis under novel lighting conditions using a traditional graphics pipeline for surface rendering.

\section{Method}
\label{sec:Method}
We will first show how to decouple appearance into reflectance and lighting, which allows to estimate a high quality material in addition to the shape.
Following this, we will extend this into a novel framework for super-resolved shape and BRDF estimation to allow rendering of novel views with more detail than the individual input images.

\subsection{Radiance field with explicit BRDF model}

We express the radiance field~$\radiance(\px,\pn,\vo)$ in~\eqref{eq:volume_integral} in terms of the BRDF and lighting using the rendering equation~\cite{kajiya1986rendering},
\begin{equation}
  \radiance(\px,\pn,\vo) = \int_{H_\pn} \radiance_i(\px,\vi) \brdf(\px, \pn, \vo, \vi) ( \vi \cdot \pn) \dx\vi,
  \label{eq:rendering_equation}
\end{equation}
where $H_\pn \subset \S^2$ is the half-sphere in direction~$\pn$, $\radiance_i(\px,\vi)$ denotes the radiance incoming at~$\px$ from direction~$\vi$,
and $\brdf$ the spatially varying BRDF (SVBRDF). %
Since we assume an achromatic point light source colocated with the camera center $\pc$, \eqref{eq:rendering_equation} simplifies to
\begin{equation}
  \radiance(\px,\pn,\vo) = \dfrac{L_0}{\norm{x - c}^2} \brdf(\px,\pn,\vo,\vo) (\vo \cdot \pn),
  \label{eq:point_light_model}
\end{equation}
where $L_0$ corresponds to the scalar light intensity.
We use use the simplified Disney BRDF \cite{karis2013real}, as this provides a compact model expressive enough to represent a wide variety of materials, which was successfully used in several prior works~\cite{luan2021unified,zhang2021physg}.
Here, the SVBRDF is parametrized with a diffuse RGB albedo~$\brdfd:\Omega\to\R^3_{\geq0}$,
a roughness $\rough:\Omega\to\R_{+}$ and a specular albedo $\spec:\Omega\to[0,1]$.

We implement these three components using two MLPs. The first MLP~$\brdfd(\px;\diffuseparam)$
is used to compute the diffuse component of the BRDF at a point $\px$,
the second MLP $\alpha(\px;\specparam) = \left(\spec(\px;\specparam),\rough(\px;\specparam)\right)$
computes the respective specular components.
The combined network parameters for BRDF are denoted with~$\brdfparam=(\diffuseparam,\specparam)$.
As mentioned earlier, we model the geometry using a third MLP~$\sdf(\px;\sdfparam)$
for the SDF with its own network parameters $\sdfparam$. 
Note that we do not incorporate a global geometric feature map into our framework.
The main motivation behind such a map is for the radiance field $\radiance$ to account for indirect lighting and self-shadows.
We follow the spirit of multiple works~\cite{bi2020deep,haefner2021recovering,luan2021unified,nam2018practical,zhang2022iron} showing that satisfactory results can be achieved without modeling indirect lighting explicitly.
Furthermore, we can successfully treat these effects as outliers thanks to the robust $L^1$-norm in~\eqref{eq:data_term}.
On the other hand, self-shadows are not present in our captures anyway, due to our colocated camera-light setup.

In order to perform inverse rendering, we train our neural networks from the available input images.
For comparing the images to the rendered image in the loss function, we model the physical camera capturing process in the next section and show how this leads to a super-resolved reconstruction of the scene.
\subsection{Super-resolution image formation model}
As common in many volume rendering based approaches~\cite{mildenhall2021nerf,yariv2021volume,wang2021neus},
it is assumed that the image brightness at a given pixel corresponds exactly to the accumulated radiance of the volume rendering~\eqref{eq:volume_integral},
\begin{equation}
  \img_\pp = \accradiance_\pp(\sdfparam,\brdfparam).
  \label{eq:image_formation}
\end{equation}
As described in \cite{wang2022nerf}, even if a framework can render novel views at any resolution during inference, the performance will significantly decrease when the inference resolution becomes larger than the one of the input images.
Indeed, during training, the networks are sampled exactly at the pixel locations of the training images, which means that there is no training data for points of the surface whose projections do not coincide with these locations.
Thanks to the interpolation property of neural networks, we can usually still compute a reasonable value for the points which were not seen during training, in particular since they are typically in between points which have been trained.
However, the sampling rate of the input images band-limits the components of the reconstructed SDF and BRDF, and higher frequency details are not magically generated if one only renders at higher resolution.

In order to restore higher frequency content, we can exploit that the camera capturing process does not only sample the exact centers of the pixels.
Instead, a camera performs an integration over a subset of incoming rays which can be modeled by the so-called point spread function (PSF).
In essence, it describes blur during the capturing process, for example caused by integration across a sensor element, diffraction, lens aberrations, or objects being not perfectly in focus~\cite{delbracio2013two}.
This intrinsic blur can actually be beneficial to avoid aliasing on the captured low resolution images, as it reduces high-frequency content in the captured scene, but leads to unavoidable loss of detail.

\begin{figure}[t]
  \centering
  \small
  \newcommand{\mywidthc}{0.02\textwidth}
  \newcommand{\mywidthx}{0.22\textwidth}
  \newcommand{\mywidthxx}{0.125\textwidth}
  \newcommand{\mywidthxxx}{0.1\textwidth}
  \newcolumntype{X}{ >{\centering\arraybackslash} m{\mywidthx} }
  \newcolumntype{C}{ >{\centering\arraybackslash} m{\mywidthc} }
  \newcommand{\tabelt}[1]{\hfil\hbox to 0pt{\hss #1 \hss}\hfil}
  \setlength\tabcolsep{1pt} %
  \begin{tabular}{CXX}
  \rotatebox{90}{Pixel}&
    \includegraphics[width=\mywidthxx]{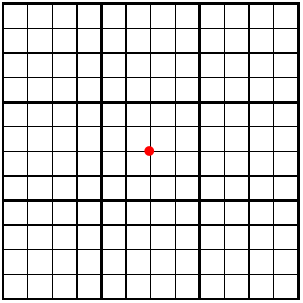}&
    \includegraphics[width=\mywidthxx]{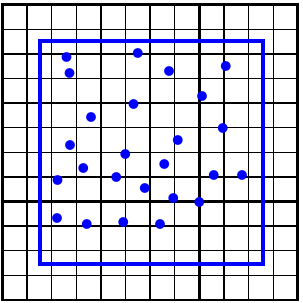}\\
  \rotatebox{90}{PSF Kernel}&
    \includegraphics[width=\mywidthxx]{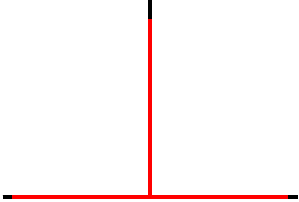}&
    \includegraphics[width=\mywidthxx]{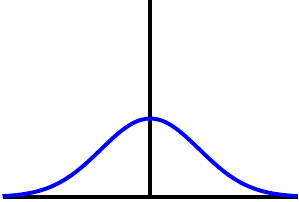}\\
    & Dirac & Gaussian
  \end{tabular}
  \caption{Visualization of standard sampling strategies~\cite{mildenhall2021nerf,yariv2021volume,zhang2021physg, zhang2022iron}, compared to our sampling strategy for a scaling factor of four (in each direction) in a $3\times3$ LR grid. Considering the low-resolution pixel in the center, a trivial Dirac kernel (bottom-left) leads to an evaluation of the network at a single (red) point (top-left). Our principled approach with a Gaussian kernel with standard deviation of half a low-resolution pixel's side length (bottom-right) naturally leads to a much denser sampling (top-right) at multiple (blue) points within a squared region (blue square) allowing for a much denser sampled network optimization.}
\label{fig:srexample}
\end{figure}

Taking this physical process into account, we follow ~\cite{baker2002limits, goldlucke2014super} and
consider the effect of the PSF in our image formation model.
Thus, generalizing~\eqref{eq:image_formation}, we convolve the accumulated radiance with the~$\PSF$ kernel
in order to obtain image irradiance,
\begin{equation}
  \img_\pp = (\accradiance(\sdfparam,\brdfparam)\ast \PSF)(\pp) = \int_{\R^2} \accradiance_{p-q}(\sdfparam,\brdfparam) \PSF(q)\dx q.
  \label{eq:psf_image_model}
\end{equation}
For the special case that the~$\PSF$ is a Dirac delta distribution, one arrives at the original model~\eqref{eq:image_formation}, so this is indeed a generalization.
In practice, we assume that the PSF is a Gaussian distribution, as~\cite{peter2001image} have shown the validity of such an approximation, and~\cite{goldlucke2014super} successfully used it to achieve texture super-resolution.
In our experiments, we choose half the size of a pixel in the low-resolution input images as the
standard deviation.
For computational efficiency, we approximate the convolution shown in~\eqref{eq:psf_image_model} by Monte Carlo integration
\begin{equation}
  \img_\pp \approx \dfrac{1}{N_s} \sum^{N_s}_{k=1}\accradiance_{p-q_k}(\sdfparam,\brdfparam),
  \label{eq:monte_carlo_psf}
\end{equation}
where $(q_k)_{k=1\hdots N_s}$ are samples drawn from the proposed PSF.
See Figure~\ref{fig:srexample} for a comparison of the sampling process based on a Dirac kernel and a Gaussian kernel.

\subsection{Final training objective}
Our final objective consists of three terms.
The first term $E_\text{RGB}$ is the data term, ensuring that the rendered images fit the input images,
\begin{equation}
  E_\text{RGB}(\sdfparam,\brdfparam) = \sum_\pp \norm{\img_\pp - \(\accradiance(\sdfparam,\brdfparam)\ast \PSF\)(\pp)}_1.
  \label{eq:data_term}
\end{equation}
Note that we use an $L^1$-norm to improve robustness against outliers.
The second term is the Eikonal term $E_\text{eik}$, which encourages $\sdf(\px;\sdfparam)$ to approximate
an SDF, this is similar as in~\cite{gropp2020implicit},
\begin{equation}
  E_\text{eik}(\sdfparam) = \sum_\px (\norm{\nabla_\px \sdf(\px;\sdfparam)} - 1 )^2.
  \label{eq:eikonal_term}
\end{equation}
Finally, we largely follow~\cite{wang2021neus} and introduce an \textit{optional} mask loss $E_\text{mask}(\sdfparam)$, allowing to impose silhouette consistency, 
\begin{equation}
  E_\text{mask}(\sdfparam) = \sum_\pp \text{BCE}(M_\pp,W_\pp(\sdfparam)),
  \label{eq:mask_term}
\end{equation}
where $M_\pp$ is the given binary mask value at the pixel $\pp$,
$W_\pp(\sdfparam) = \sum_{i=1}^{m-1} \pdf(t_i;\sdfparam)$
is the sum of the weights at the sampling locations~$t_i$ used in~\eqref{eq:discrete_volume_integral}, and BCE is the binary cross entropy loss~\cite{wang2021neus}.
We would like to emphasize that the only aim of the mask loss is to reduce computation time as it allows us to use more samples per pixel inside the mask, and one single sample per pixel outside.
Additionally, if the mask loss is used, we also truncate the volume rendering integral~\eqref{eq:volume_integral} to the unit sphere, as the background is already modeled by the mask loss, further reducing the computational requirements.
If the mask loss is not used, we use an inverse sphere parameterization for the background~\cite{zhang2020nerf++}.
Both use cases are directly inspired from~\cite{wang2021neus}.

The final loss becomes the sum of the three terms with additional weighting parameters $\lambda_1, \lambda_2\geq0$,
\begin{equation}
  E(\sdfparam,\brdfparam) = E_\text{RGB}(\sdfparam,\brdfparam) + \lambda_1 E_\text{eik}(\sdfparam) + \lambda_2 E_\text{mask}(\sdfparam).
  \label{eq:loss_function}
\end{equation}
This loss function can in principle be used to estimate a super-resolved geometry and BRDF with one single training pass.
However, since the full super-resolution model is computationally expensive, we first run an initialization pass
with a Dirac kernel for the PSF and without the mask loss.
Indeed, in Section~\ref{sec:Results} we show that this super-resolution free approach
already retrieves state-of-the-art results even without mask supervision.
After that, we have accurate silhouettes from projecting the estimated geometry into the input images,
and can significantly decrease computation time by optimizing sampling as described above.
Furthermore, convergence of the full super-resolution model is accelerated since we already start from a reasonable initialization of the scene.

\section{Results}
\label{sec:Results}
We evaluate our framework \supervol{} on both synthetic and real world data.
To this end, we create a synthetic dataset which consists of a combination of two geometries and two materials, designated in the following as \dogA{}, \dogB{}, \girlA{} and \girlB{}. Each of them is rendered from 60 different viewpoints for training. 30 other viewpoints are rendered for testing, including non-colocated illumination to evaluate our approach's generalization capability. Our real world dataset consist of four scans: \bird{} and \squirrel{}, which we captured ourselves, and \pony{} and \dragon{} from \cite{bi2020deep}. From these, we consider between 30 and 60 images for training and around 30 images for testing.
\paragraph{Evaluation.}
\begin{table*}[t]
\footnotesize
\setlength{\tabcolsep}{1pt}
\begin{tabular*}{\textwidth}{l||cc|cc|cc||ccc|ccc|ccc}
~ & \multicolumn{2}{c|}{$\uparrow$PSNR} & \multicolumn{2}{c|}{$\uparrow$SSIM~\cite{1284395}} & \multicolumn{2}{c||}{$\downarrow$MAE} & \multicolumn{3}{c|}{$\uparrow$PSNR} & \multicolumn{3}{c|}{$\uparrow$SSIM~\cite{1284395}} & \multicolumn{3}{c}{$\downarrow$MAE} \\
~ & \cite{zhang2022iron} & \nosr & \cite{zhang2022iron} & \nosr & \cite{zhang2022iron} & \nosr & \cite{zhang2022iron} & \nosr & \supervol & \cite{zhang2022iron} & \nosr & \supervol & \cite{zhang2022iron} & \nosr & \supervol\\
\hline
synthetic & 30.4949 & \textbf{34.7325} & 0.9095 & \textbf{0.9508} & 7.8286 & \textbf{4.8109} & 28.9161 & 31.2778 & \textbf{32.1521} & 0.8643 & 0.8973 & \textbf{0.9137} & 9.3410 & 5.9897 & \textbf{5.5785}\\
synthetic non-colocated & 31.7933 & \textbf{35.8004} & 0.9026 & \textbf{0.9475} & 7.8286 & \textbf{4.8109} & 30.4858 & 33.3652 & \textbf{33.7498} & 0.8695 & 0.9085 & \textbf{0.9208} & 9.3410 & 5.9897 & \textbf{5.5785}\\
real world & 29.1411 & \textbf{32.2888} & 0.8721 & \textbf{0.9276} & $\times$ & $\times$ & 28.9149 & 31.5179 & \textbf{31.7068} & 0.8817 & 0.9143 & \textbf{0.9157} & $\times$ & $\times$ & $\times$\\\multicolumn{16}{c}{}\\[-0.5em]
\multicolumn{1}{c}{} & \multicolumn{6}{c}{\small{(a) high resolution training}} & \multicolumn{9}{c}{\small{(b) low resolution training}}\\[0.5em]
\end{tabular*}
\caption{
Average PSNR, SSIM~\cite{1284395} and MAE across the datasets using high and the more challenging low resolution input for training, respectively.
In both training scenarios, \supervol{} and  its simplified counterpart \nosr{} (with a Dirac kernel) outperform IRON~\cite{zhang2022iron} quantitatively.
Geometric reconstruction and image synthesis quality is in favor of our approach for the tested novel and unseen viewpoints.
Additionally, our method can generalize well to a non-colocated lighting setup.
Note, \supervol{} and \nosr{} trained on low resolution images can perform better than IRON~\cite{zhang2022iron} trained on high resolution input.
}
\label{tab:training}
\end{table*}
We consider an ablation study that we name "noSR", which consists of our framework SupeRVol, but with a Dirac kernel instead of a Gaussian kernel, and we evaluate both noSR and SupeRVol together with IRON\cite{zhang2022iron}. IRON has demonstrated state-of-the-art performance for inverse rendering with photometric images, significantly outperforming prior work~\cite{bi2020deep,luan2021unified} for which no open-source code is available online.
Further, we consider two complementary scenarios.
In {\em high resolution training}, we use the original (high resolution) images of the scans for both training and testing.
In {\em low resolution training}, we down-sample the resolution of the training images of the scans by a factor of four %
to mimic the effect of a low-resolution image capture, and then test on the high resolution test images.

The first scenario is typically considered in inverse rendering works, and allows to assess the quality of our differentiable volume renderer with an explicit BRDF model.
Since both, training and testing resolutions are the same in this case, we only compare \nosr{} against IRON.
For the second scenario, we include \supervol{} in the comparison, allowing to measure the impact of modeling the PSF%
.

\paragraph{High resolution training.}
\begin{figure}[t]
  \centering
  \small
  \newcommand{\mywidthx}{0.145\textwidth}
  \newcolumntype{Y}{ >{\centering\arraybackslash} m{\mywidthx} }
  \newcommand{\tabelt}[1]{\hfil\hbox to 0pt{\hss #1 \hss}\hfil}
  \setlength\tabcolsep{1pt} %
  \begin{tabular}{cYYY}
    \rotatebox[origin=c]{90}{\bird{}}&
    \includegraphics[width=\mywidthx]{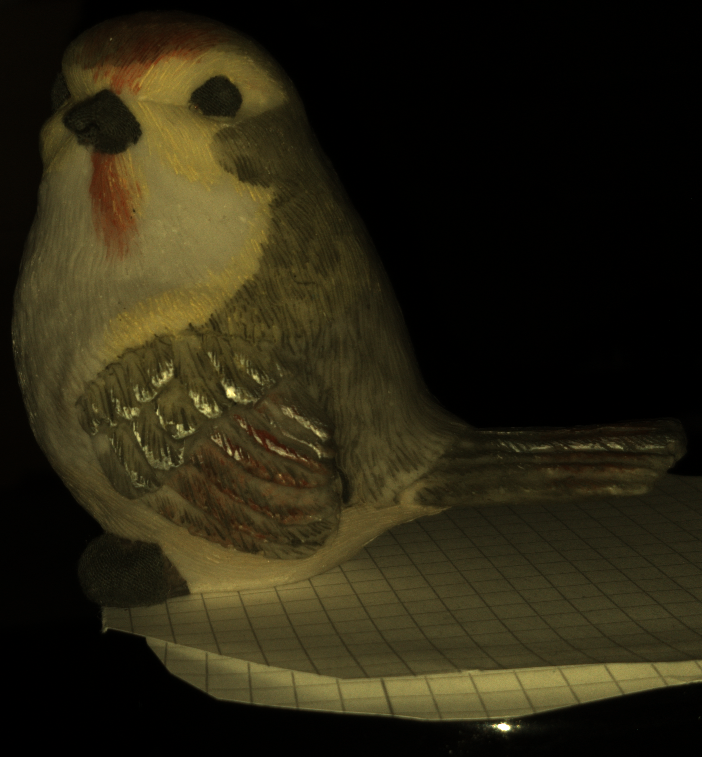}&
    \includegraphics[width=\mywidthx]{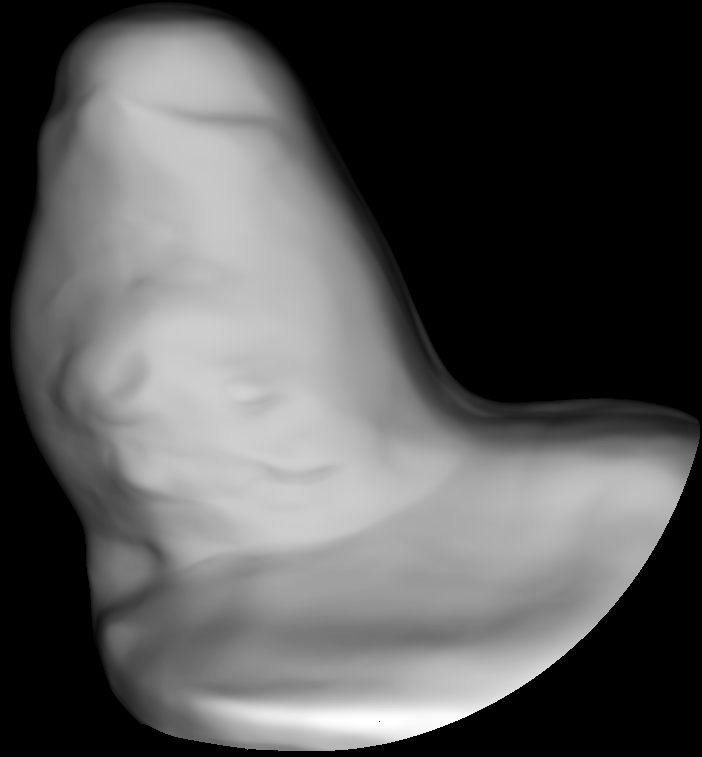}&
    \includegraphics[width=\mywidthx]{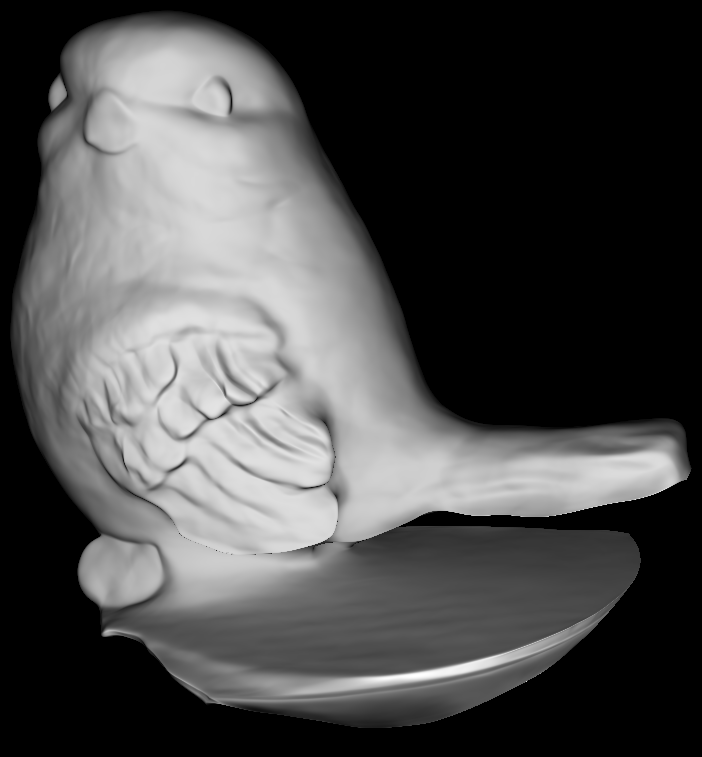}\\
    \rotatebox[origin=c]{90}{\pony{}}&
    \includegraphics[width=\mywidthx]{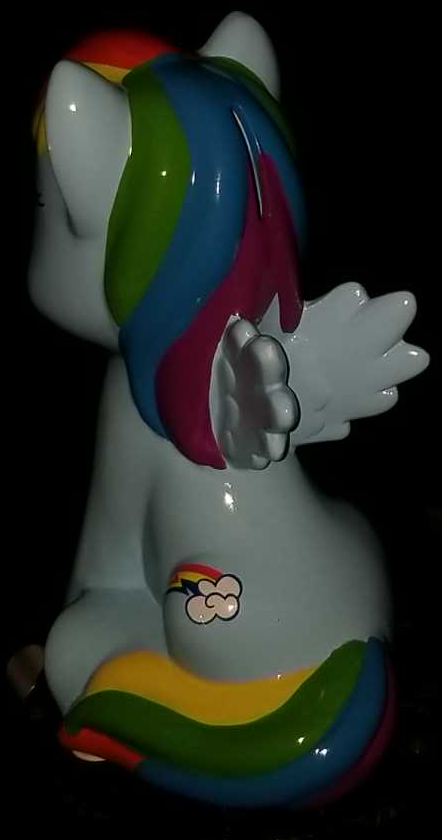}&
    \includegraphics[width=\mywidthx]{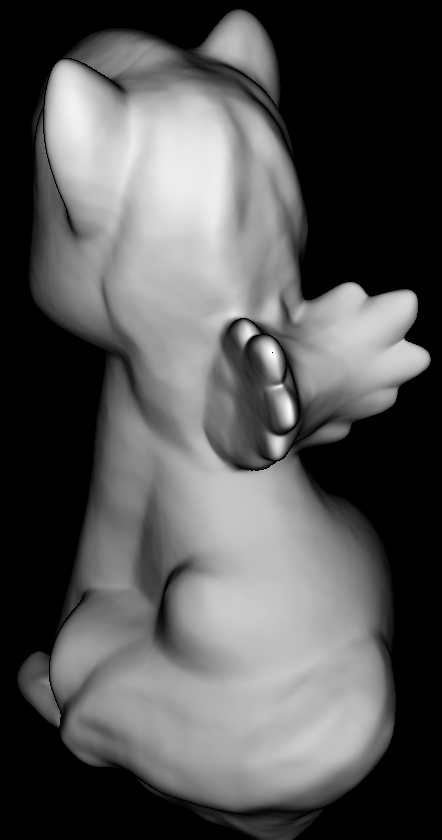}&
    \includegraphics[width=\mywidthx]{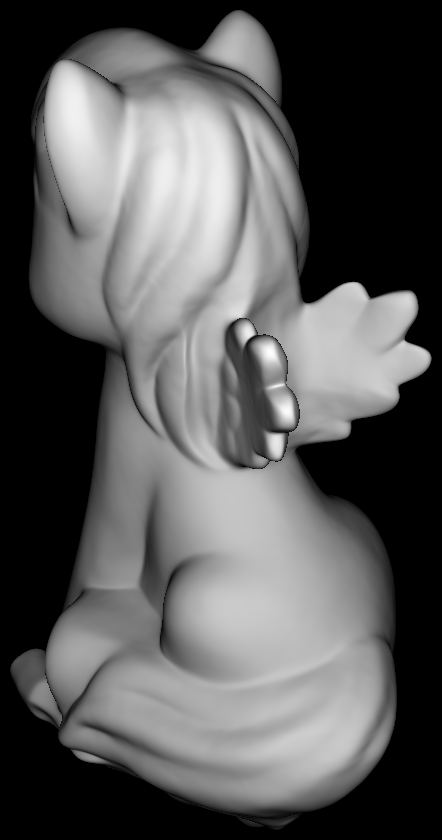}\\
    &Real Image & IRON\cite{zhang2022iron} & \nosr{}
  \end{tabular}
  \caption{Estimated geometry after high resolution training. IRON\cite{zhang2022iron} produces significant artifacts at concave parts.
  Additionally, it lacks of detail in convex regions.
  Instead, with its inverse volume rendering, \nosr{} produces sharp and artifact free reconstructions in both regions of the objects. 
  }
\label{fig:HR_concave_parts}
\end{figure}

\begin{figure}[t]
  \centering
  \small
  \newcommand{\mywidthx}{0.145\textwidth}
  \newcolumntype{Y}{ >{\centering\arraybackslash} m{\mywidthx} }
  \newcommand{\tabelt}[1]{\hfil\hbox to 0pt{\hss #1 \hss}\hfil}
  \setlength\tabcolsep{1pt} %
  \begin{tabular}{cYYYY}
    \rotatebox[origin=c]{90}{\girlB}
    &
    \begin{tikzpicture}[spy using outlines={rectangle,connect spies}]
      \node[anchor=south west,inner sep=0]  at (0,0) {\includegraphics[width=\mywidthx]{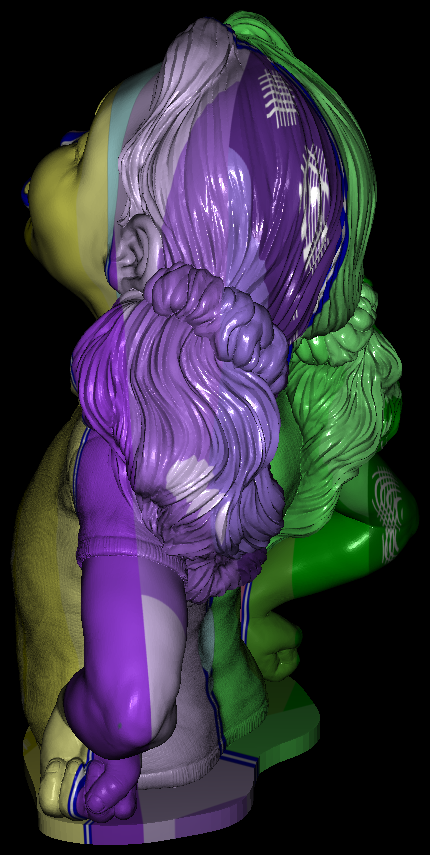}};
      \spy[color=green,width=0.5cm,height=1.5cm, magnification=2] on (1.85,3.8) in node [right] at (2.03,0.76);
    \end{tikzpicture}
    &
    \begin{tikzpicture}[spy using outlines={rectangle,connect spies}]
      \node[anchor=south west,inner sep=0]  at (0,0) {\includegraphics[width=\mywidthx]{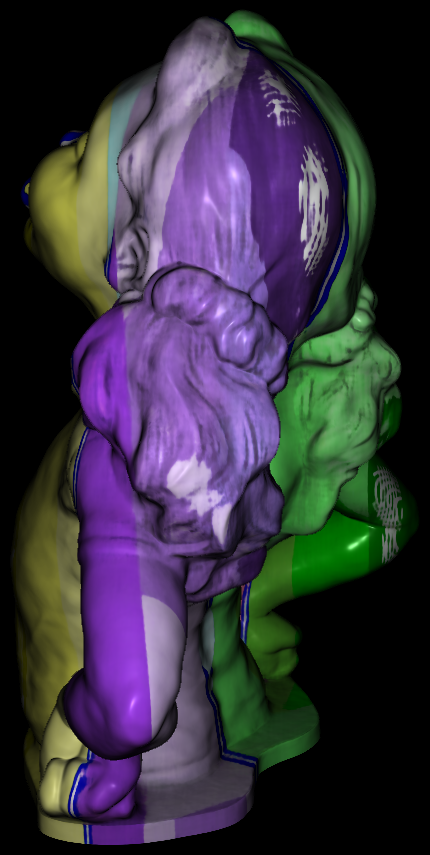}};
      \spy[color=green,width=0.5cm,height=1.5cm, magnification=2] on (1.85,3.8) in node [right] at (2.03,0.76);
    \end{tikzpicture}
    &
    \begin{tikzpicture}[spy using outlines={rectangle,connect spies}]
      \node[anchor=south west,inner sep=0]  at (0,0) {\includegraphics[width=\mywidthx]{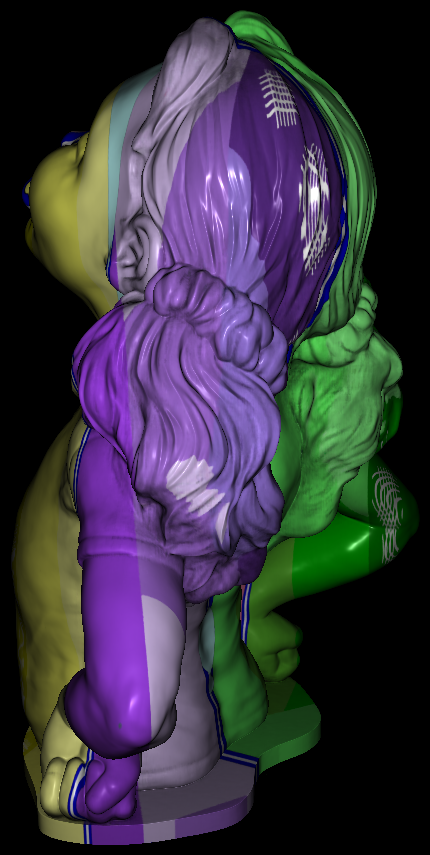}};
      \spy[color=green,width=0.5cm,height=1.5cm, magnification=2] on (1.85,3.8) in node [right] at (2.03,0.76);
    \end{tikzpicture}
    \\
    \rotatebox[origin=c]{90}{\dragon}
    &
    \begin{tikzpicture}[spy using outlines={rectangle,connect spies}]
      \node[anchor=south west,inner sep=0]  at (0,0) {\includegraphics[width=\mywidthx]{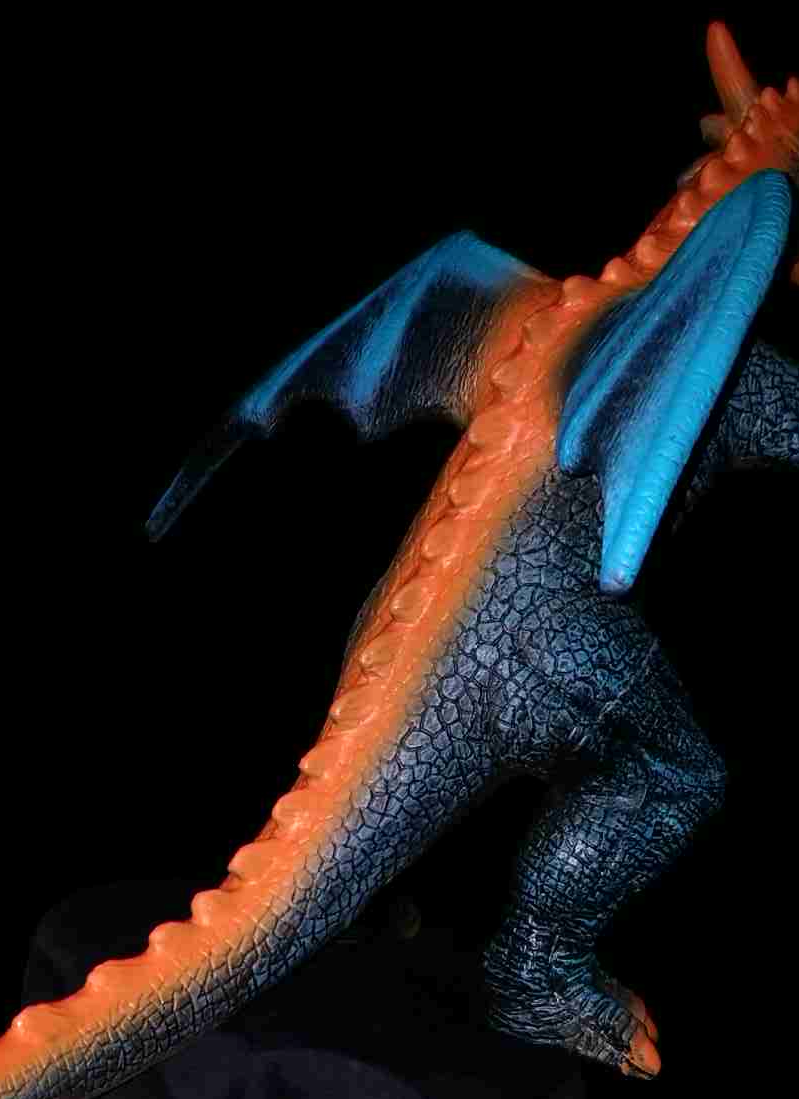}};
      \spy[color=green,width=1.0cm,height=1.0cm,magnification=2] on (1.6,1.65) in node [right] at (0.01,2.99);
    \end{tikzpicture}
    &
    \begin{tikzpicture}[spy using outlines={rectangle,connect spies}]
      \node[anchor=south west,inner sep=0]  at (0,0) {\includegraphics[width=\mywidthx]{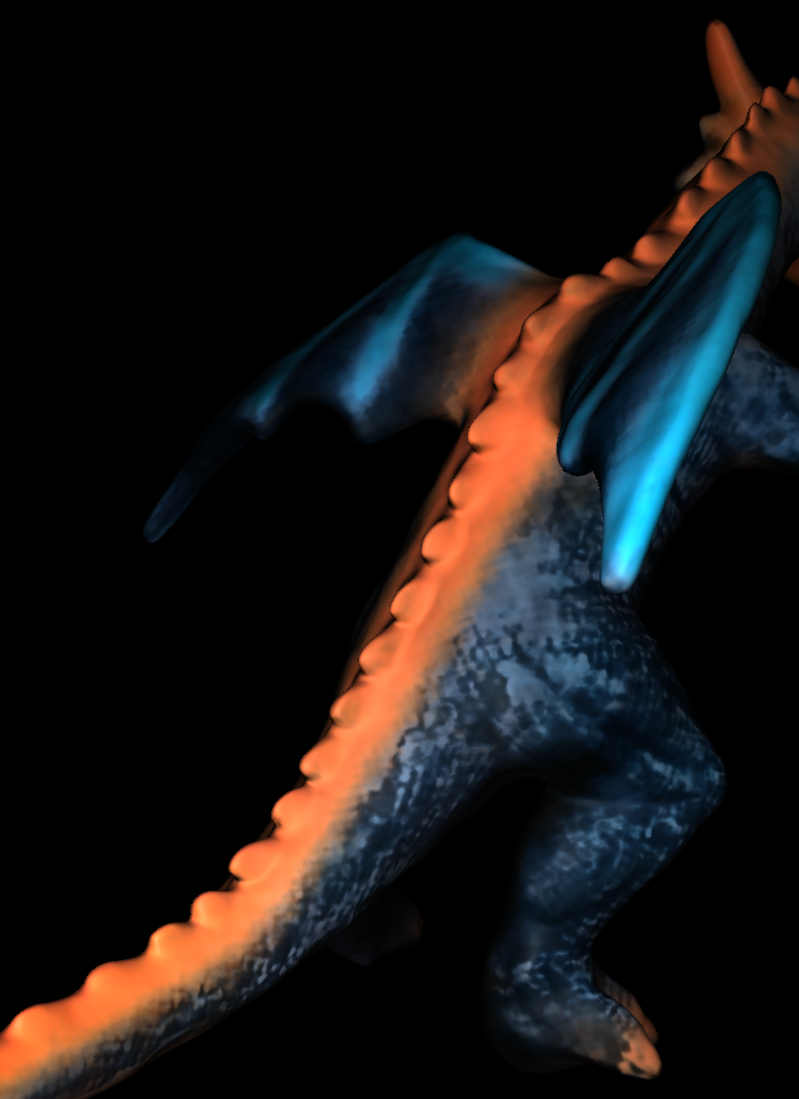}};
      \spy[color=green,width=1.0cm,height=1.0cm,magnification=2] on (1.6,1.65) in node [right] at (0.01,2.99);
    \end{tikzpicture}
    &
    \begin{tikzpicture}[spy using outlines={rectangle,connect spies}]
      \node[anchor=south west,inner sep=0]  at (0,0) {\includegraphics[width=\mywidthx]{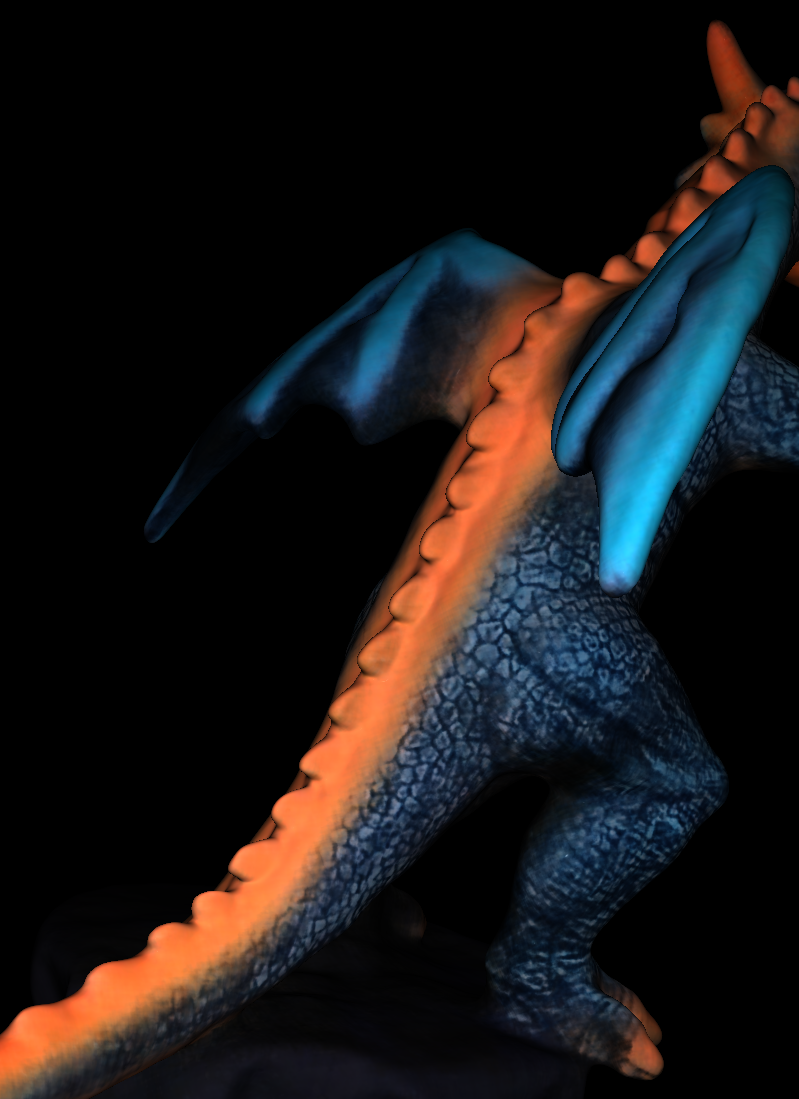}};
      \spy[color=green,width=1.0cm,height=1.0cm,magnification=2] on (1.6,1.65) in node [right] at (0.01,2.99);
    \end{tikzpicture}
    \\
    \rotatebox[origin=c]{90}{\squirrel}
    &
    \begin{tikzpicture}[spy using outlines={rectangle,connect spies}]
      \node[anchor=south west,inner sep=0]  at (0,0) {\includegraphics[width=\mywidthx]{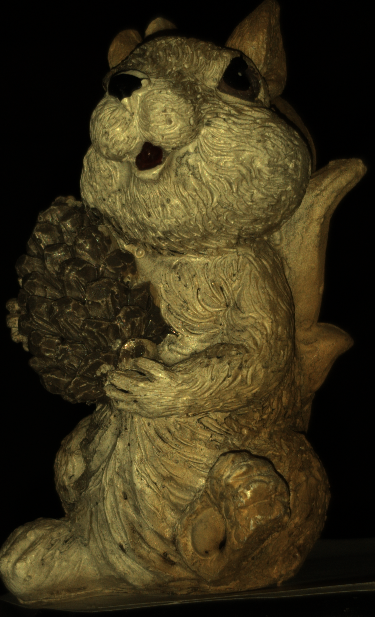}};
      \spy[color=green,width=0.5cm,height=1.0cm,magnification=2] on (1.6,2.0) in node [right] at (0.01,3.65);
    \end{tikzpicture}
    &
    \begin{tikzpicture}[spy using outlines={rectangle,connect spies}]
      \node[anchor=south west,inner sep=0]  at (0,0) {\includegraphics[width=\mywidthx]{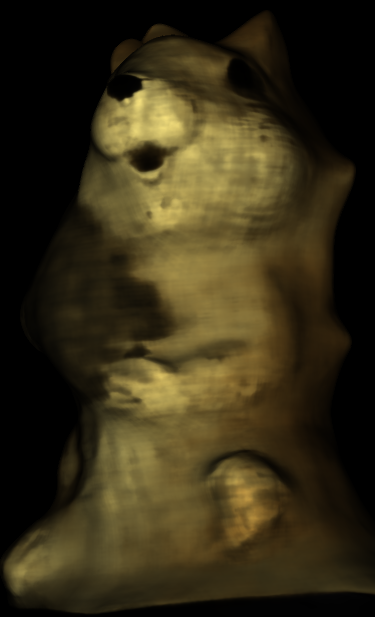}};
      \spy[color=green,width=0.5cm,height=1.0cm,magnification=2] on (1.6,2.0) in node [right] at (0.01,3.65);
    \end{tikzpicture}
    &
    \begin{tikzpicture}[spy using outlines={rectangle,connect spies}]
      \node[anchor=south west,inner sep=0]  at (0,0) {\includegraphics[width=\mywidthx]{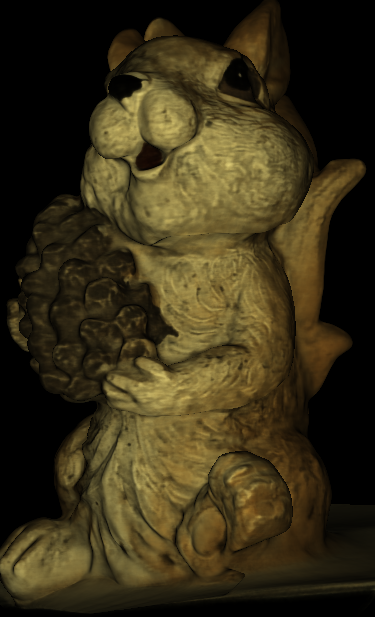}};
      \spy[color=green,width=0.5cm,height=1.0cm,magnification=2] on (1.6,2.0) in node [right] at (0.01,3.65);
    \end{tikzpicture}
    \\
    & {\small real image}& {\small IRON\cite{zhang2022iron}} & {\small noSR}    
  \end{tabular}
  \caption{Image synthesis results of novel viewpoints with a co-located light source after high resolution training.
  Our simplified approach \nosr{} results in a significantly sharper reconstruction, accurately reproducing fine scale details that IRON\cite{zhang2022iron} fails to recover.
  Notice the loss of detail in the hair of \girlB{}.
  The \dragon{}'s scales are partially lost.
  The \squirrel{} seems to be overall blurry.
  \nosr{} reconstructs sharper images across all datasets.
  }
\label{fig:HR_rendering}
\end{figure}

As can be seen in Table~\ref{tab:training} (a), \nosr{} is quantitatively superior to IRON~\cite{zhang2022iron} in both cases, i.e. geometric quality and image synthesis.
Qualitatively, geometric estimates are shown in Figure~\ref{fig:HR_concave_parts}.
It can be seen that \nosr{}
yields more detailed shape reconstruction in convex parts of the objects, e.g. the beak, eyes, and wings of \bird{} or the hair, and tail of \pony{}.
Additionally, thanks to the volume rendering based approach, it is also more reliable in highly concave parts, e.g. the tail of \bird{} or the wings of \pony{}.
IRON~\cite{zhang2022iron}, as a surface rendering based approach, fails to accurately recover these geometric details and gets stuck in local optima.
This further exhibits the advantage of using volume rendering instead of surface rendering for a better convergence~\cite{yariv2021volume,wang2021neus}, see Figure~\ref{fig:volvssurf}.
Image synthesis quality is visualized in Figure~\ref{fig:HR_rendering}, where \nosr{} demonstrates more crisp results compared to the blurred reconstructions of IRON~\cite{zhang2022iron}.
Our evaluation reveals a better inverse rendering performance when using our %
\nosr{} approach even without properly modeling the PSF.

\begin{figure*}[t]
  \centering
  \small
  \newcommand{\mywidthx}{0.205\textwidth}
  \newcommand{\mywidthy}{0.13\textwidth}
  \newcommand{\mywidthz}{0.07\textwidth}
  \newcolumntype{Y}{ >{\centering\arraybackslash} m{\mywidthx} }
  \newcommand{\tabelt}[1]{\hfil\hbox to 0pt{\hss #1 \hss}\hfil}
  \setlength\tabcolsep{1pt} %
  \begin{tabular}{cYYYY}
    & Real Image & IRON\cite{zhang2022iron} & noSR & SupeRVol
    \\
    \rotatebox[origin=c]{90}{\dogA}
    &
    \begin{tikzpicture}[spy using outlines={rectangle,connect spies}]
      \node[anchor=south west,inner sep=0]  at (0,0) {\includegraphics[width=\mywidthx]{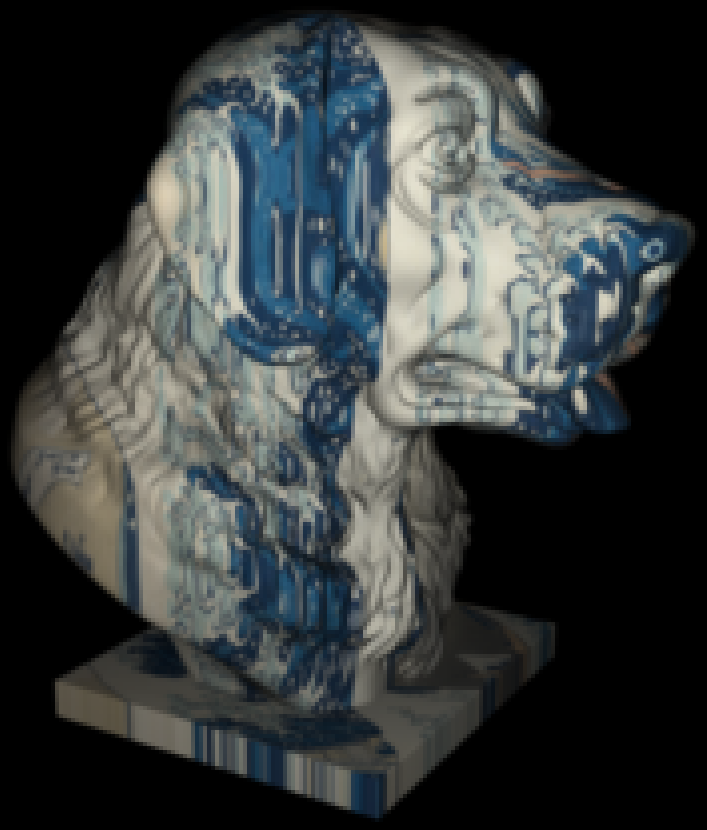}};
      \spy[color=green,width=1.5cm,height=0.9cm, magnification=2] on (1.4,0.625) in node [right] at (2.08,0.47);
    \end{tikzpicture}
    &
    \begin{tikzpicture}[spy using outlines={rectangle,connect spies}]
      \node[anchor=south west,inner sep=0]  at (0,0) {\includegraphics[width=\mywidthx]{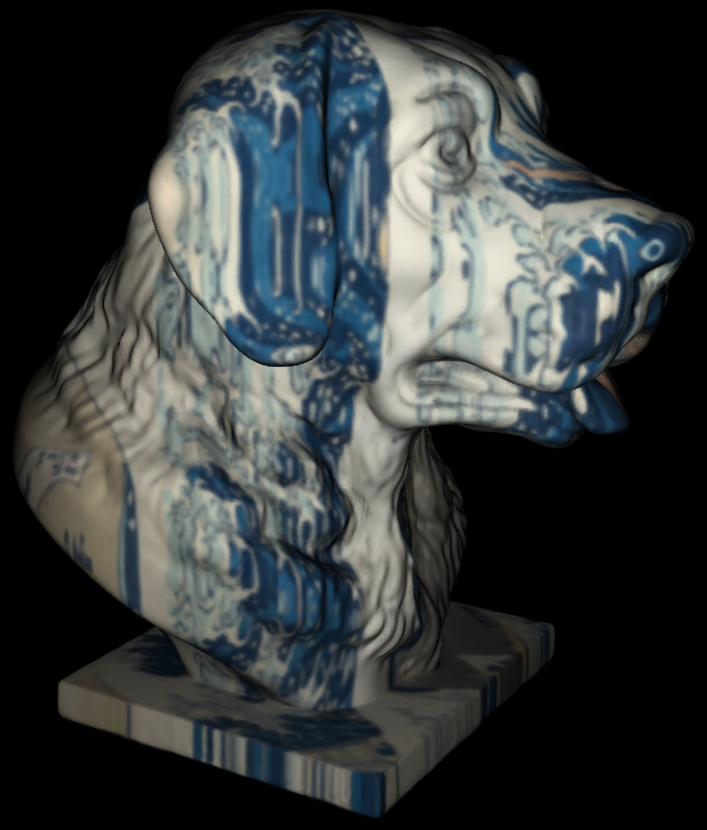}};
      \spy[color=green,width=1.5cm,height=0.9cm, magnification=2] on (1.4,0.625) in node [right] at (2.08,0.47);
    \end{tikzpicture}
    &
    \begin{tikzpicture}[spy using outlines={rectangle,connect spies}]
      \node[anchor=south west,inner sep=0]  at (0,0) {\includegraphics[width=\mywidthx]{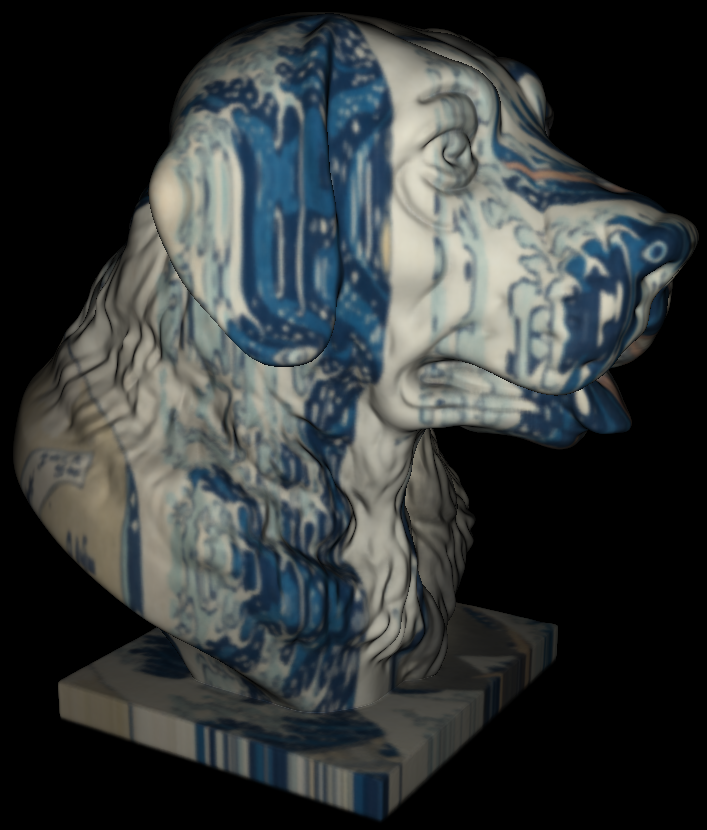}};
      \spy[color=green,width=1.5cm,height=0.9cm, magnification=2] on (1.4,0.625) in node [right] at (2.08,0.47);
    \end{tikzpicture}
    &
    \begin{tikzpicture}[spy using outlines={rectangle,connect spies}]
      \node[anchor=south west,inner sep=0]  at (0,0) {\includegraphics[width=\mywidthx]{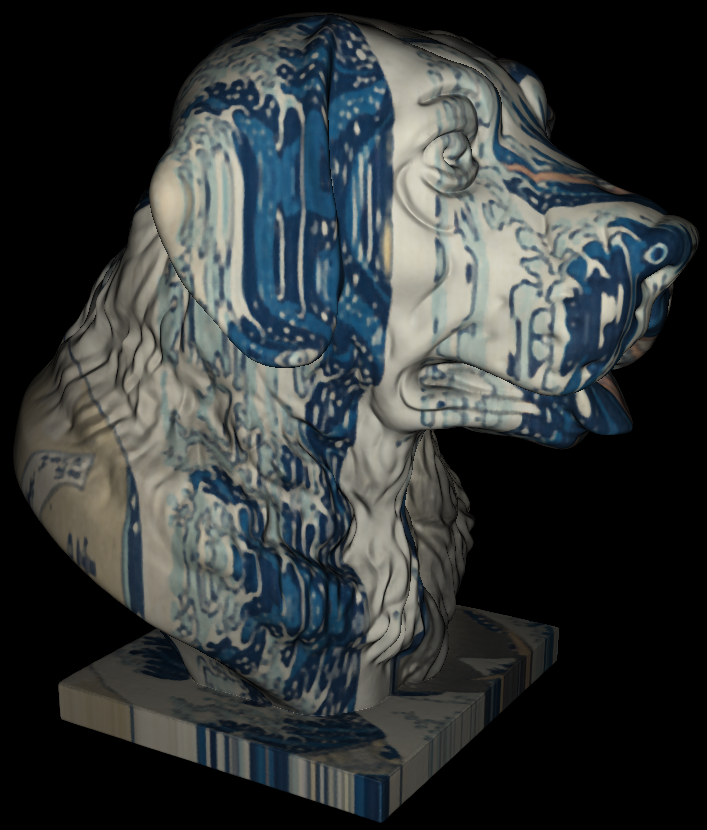}};
      \spy[color=green,width=1.5cm,height=0.9cm, magnification=2] on (1.4,0.625) in node [right] at (2.08,0.47);
    \end{tikzpicture}
    \\
    \rotatebox[origin=c]{90}{\pony}
    &
    \begin{tikzpicture}[spy using outlines={rectangle,connect spies}]
      \node[anchor=south west,inner sep=0]  at (0,0) {\includegraphics[width=\mywidthx]{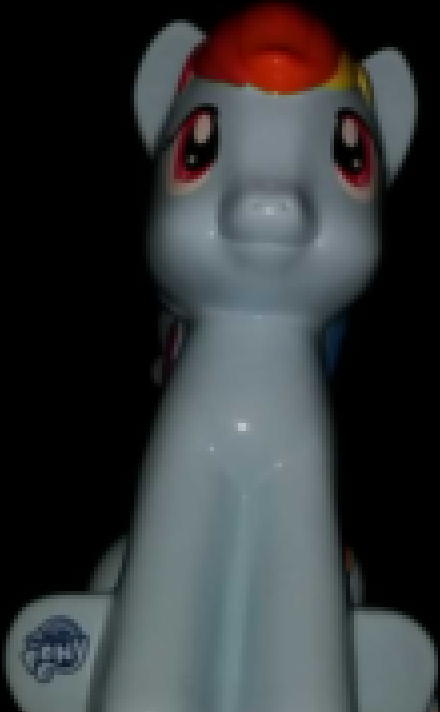}};
      \spy[color=green,width=1.2cm,height=1.2cm, magnification=2] on (0.45,0.51) in node [right] at (0.01,3);
    \end{tikzpicture}&
    \begin{tikzpicture}[spy using outlines={rectangle,connect spies}]
      \node[anchor=south west,inner sep=0]  at (0,0) {\includegraphics[width=\mywidthx]{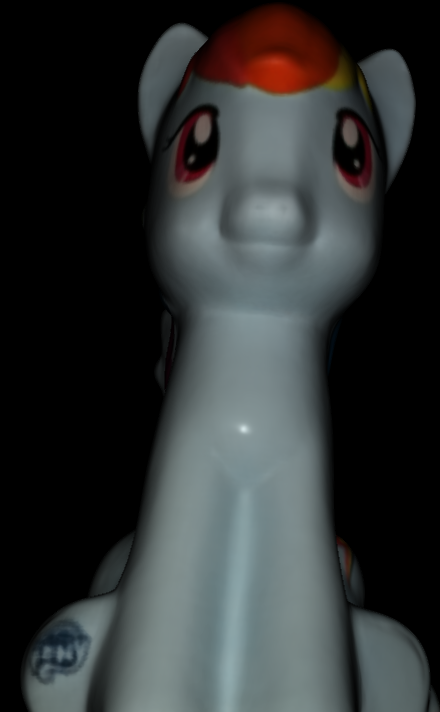}};
      \spy[color=green,width=1.2cm,height=1.2cm, magnification=2] on (0.45,0.51) in node [right] at (0.01,3);
    \end{tikzpicture}&
    \begin{tikzpicture}[spy using outlines={rectangle,connect spies}]
      \node[anchor=south west,inner sep=0]  at (0,0) {\includegraphics[width=\mywidthx]{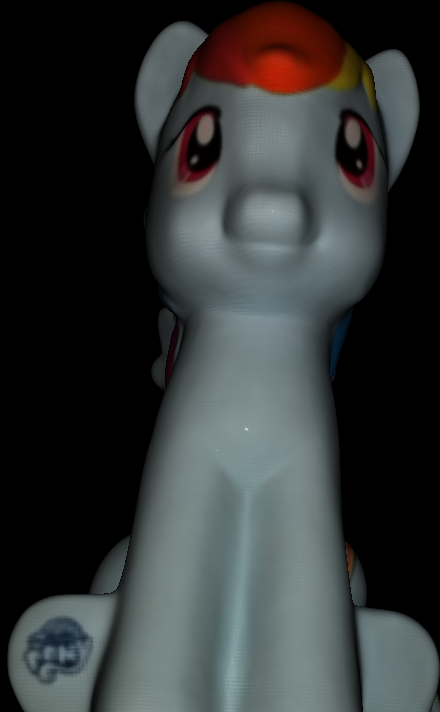}};
      \spy[color=green,width=1.2cm,height=1.2cm, magnification=2] on (0.45,0.51) in node [right] at (0.01,3);
    \end{tikzpicture}&
    \begin{tikzpicture}[spy using outlines={rectangle,connect spies}]
      \node[anchor=south west,inner sep=0]  at (0,0) {\includegraphics[width=\mywidthx]{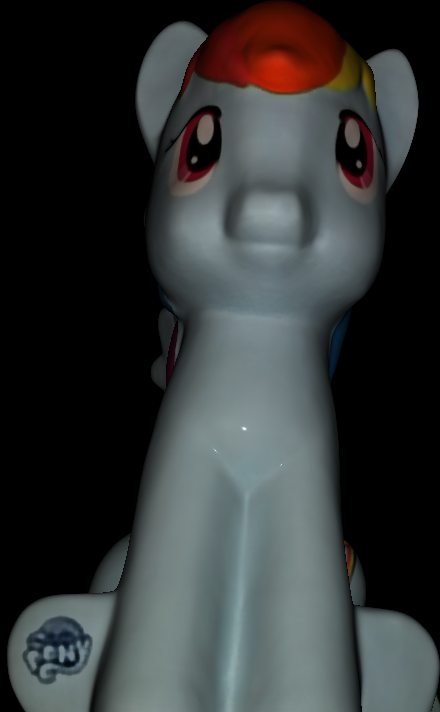}};
      \spy[color=green,width=1.2cm,height=1.2cm, magnification=2] on (0.45,0.51) in node [right] at (0.01,3);
    \end{tikzpicture}\\
    \rotatebox[origin=c]{90}{\bird}
    &
    \begin{tikzpicture}[spy using outlines={rectangle,connect spies}]
      \node[anchor=south west,inner sep=0]  at (0,0) {\includegraphics[width=\mywidthx]{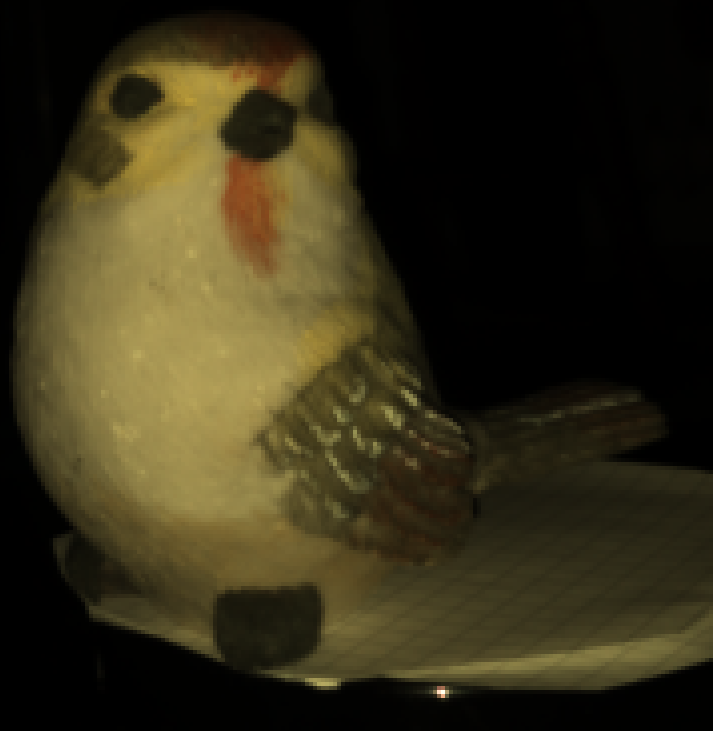}};
      \spy[color=green,width=1.2cm,height=1.2cm, magnification=5] on (1.6,1.555) in node [right] at (2.39,3.075);
    \end{tikzpicture}&
    \begin{tikzpicture}[spy using outlines={rectangle,connect spies}]
      \node[anchor=south west,inner sep=0]  at (0,0) {\includegraphics[width=\mywidthx]{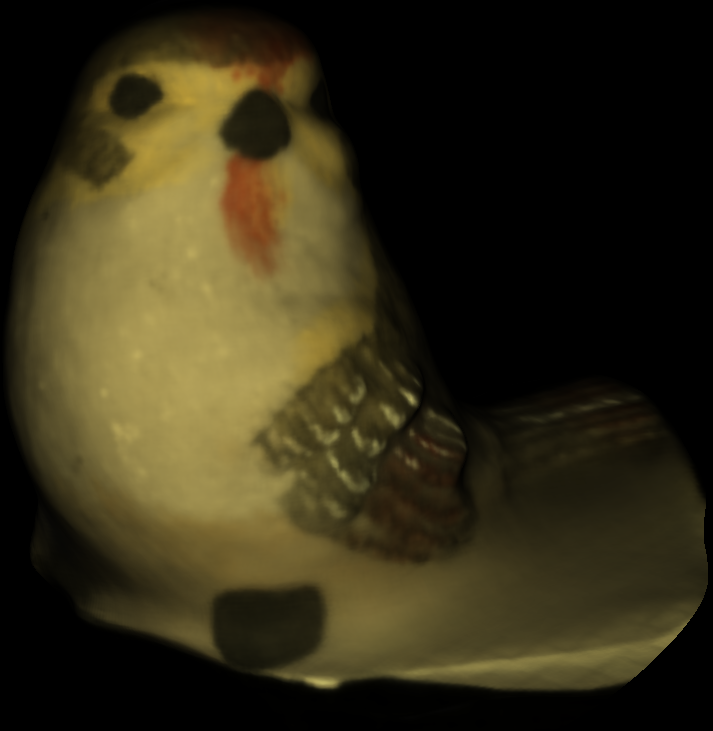}};
      \spy[color=green,width=1.2cm,height=1.2cm, magnification=5] on (1.6,1.555) in node [right] at (2.39,3.075);
    \end{tikzpicture}&
    \begin{tikzpicture}[spy using outlines={rectangle,connect spies}]
      \node[anchor=south west,inner sep=0]  at (0,0) {\includegraphics[width=\mywidthx]{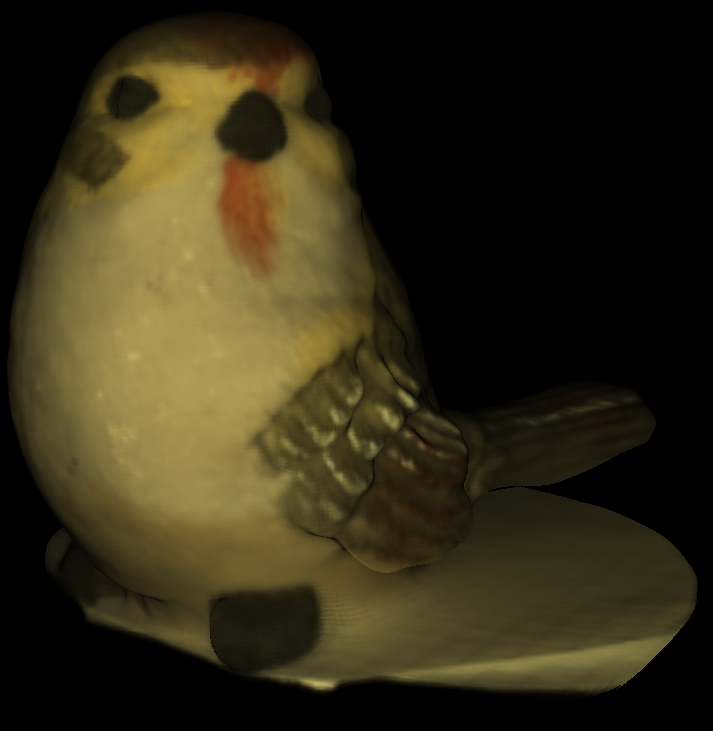}};
      \spy[color=green,width=1.2cm,height=1.2cm, magnification=5] on (1.6,1.555) in node [right] at (2.39,3.075);
    \end{tikzpicture}&
    \begin{tikzpicture}[spy using outlines={rectangle,connect spies}]
      \node[anchor=south west,inner sep=0]  at (0,0) {\includegraphics[width=\mywidthx]{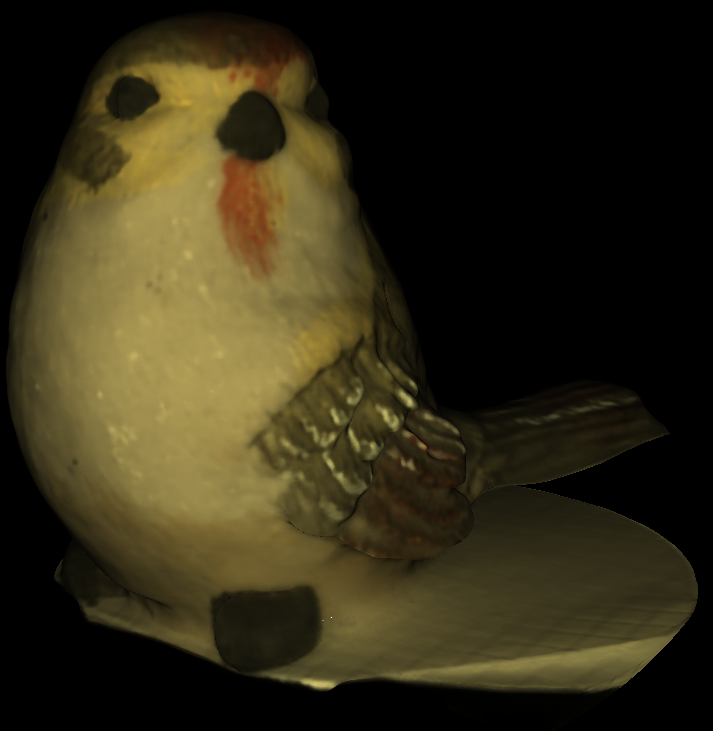}};
      \spy[color=green,width=1.2cm,height=1.2cm, magnification=5] on (1.6,1.555) in node [right] at (2.39,3.075);
    \end{tikzpicture}
  \end{tabular}
  \caption{
  Image synthesis results of novel viewpoints with a colocated light source after low resolution training.
  The rendering obtained with our complete \supervol{} model is much sharper, with unprecedented details that are lost in the remaining approaches.
  }
\label{fig:LR_rendering}
\end{figure*}

\begin{figure*}[t]
  \small
  \newcommand{\mywidthx}{0.185\textwidth}
  \newcommand{\mywidthy}{0.08\textwidth}
  \newcolumntype{Y}{ >{\centering\arraybackslash} m{\mywidthx} }
  \newcommand{\tabelt}[1]{\hfil\hbox to 0pt{\hss #1 \hss}\hfil}
  \setlength\tabcolsep{1pt} %
  \centerline{
  \begin{tabular}{cYYYYY}
    \rotatebox[origin=c]{90}{\girlA{}}&
    \begin{tikzpicture}[spy using outlines={rectangle,connect spies}]
      \node[anchor=south west,inner sep=0]  at (0,0) {    \includegraphics[width=\mywidthx]{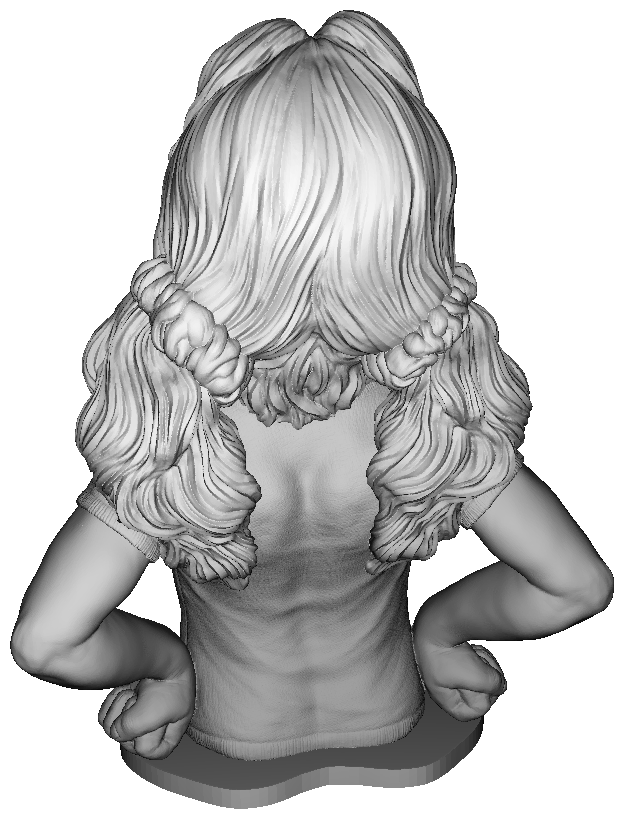}};
      \spy[color=green,width=1.3cm,height=1.3cm, magnification=5] on (2.15,2.4) in node [right] at (-0.6,3.55);
    \end{tikzpicture}&
    \begin{tikzpicture}[spy using outlines={rectangle,connect spies}]
      \node[anchor=south west,inner sep=0]  at (0,0) {    \includegraphics[width=\mywidthx]{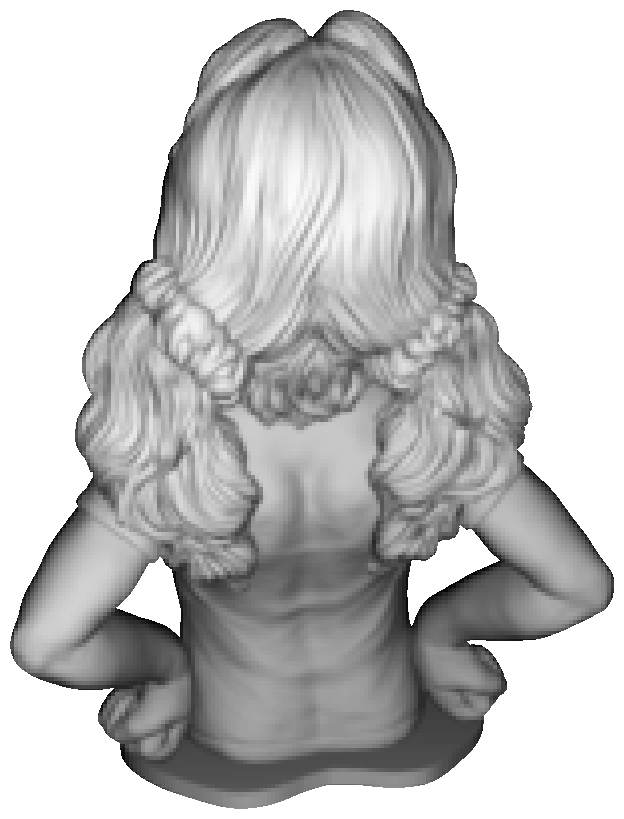}};
      \spy[color=green,width=1.3cm,height=1.3cm, magnification=5] on (2.15,2.4) in node [right] at (-0.6,3.55);
    \end{tikzpicture}&
    \begin{tikzpicture}[spy using outlines={rectangle,connect spies}]
      \node[anchor=south west,inner sep=0]  at (0,0) {    \includegraphics[width=\mywidthx]{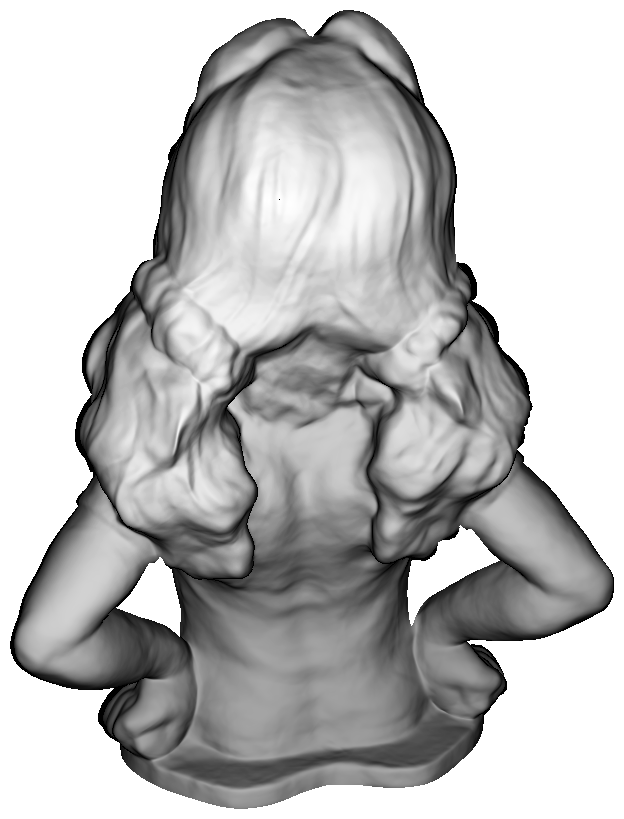}};
      \spy[color=green,width=1.3cm,height=1.3cm, magnification=5] on (2.15,2.4) in node [right] at (-0.6,3.55);
    \end{tikzpicture}&
    \begin{tikzpicture}[spy using outlines={rectangle,connect spies}]
      \node[anchor=south west,inner sep=0]  at (0,0) {    \includegraphics[width=\mywidthx]{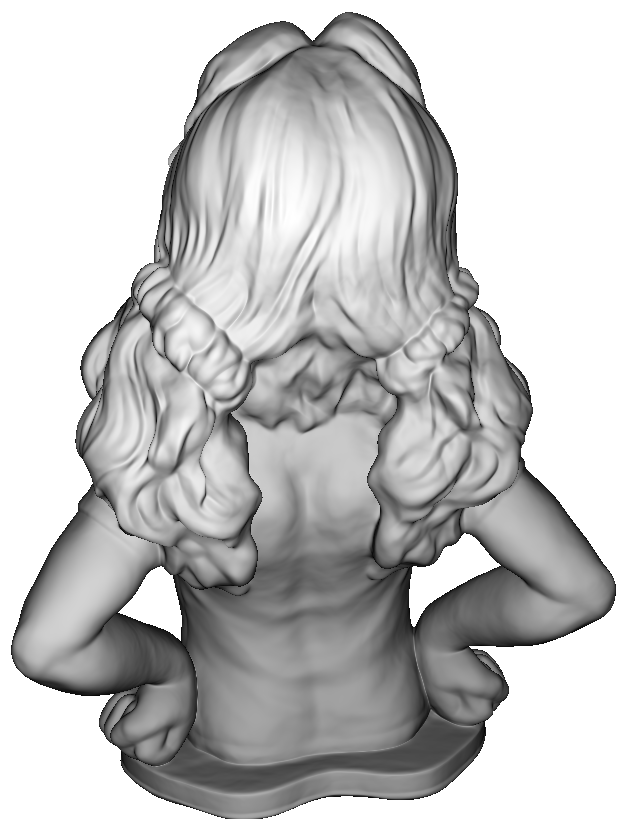}};
      \spy[color=green,width=1.3cm,height=1.3cm, magnification=5] on (2.15,2.4) in node [right] at (-0.6,3.55);
    \end{tikzpicture}&
    \begin{tikzpicture}[spy using outlines={rectangle,connect spies}]
      \node[anchor=south west,inner sep=0]  at (0,0) {    \includegraphics[width=\mywidthx]{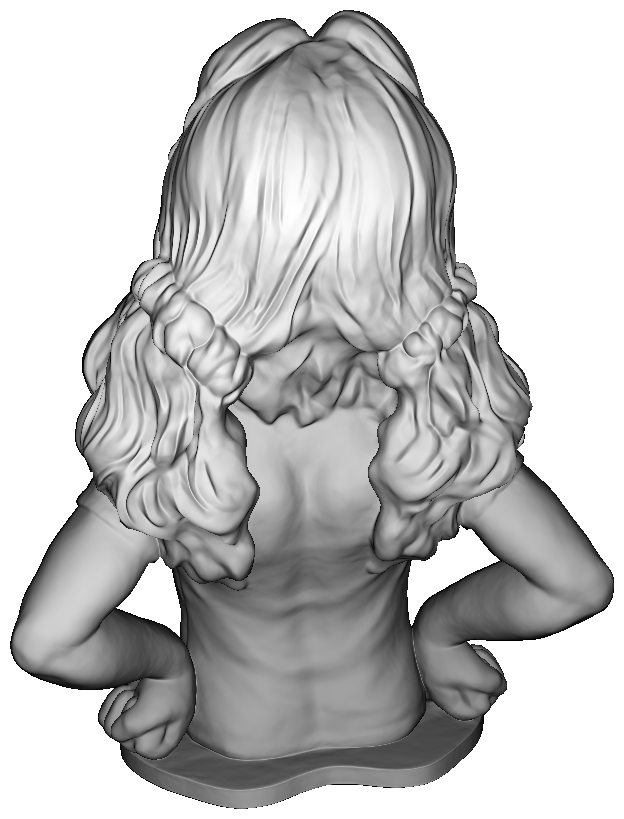}};
      \spy[color=green,width=1.3cm,height=1.3cm, magnification=5] on (2.15,2.4) in node [right] at (-0.6,3.55);
    \end{tikzpicture}\\[1mm]
    &HR & LR & IRON\cite{zhang2022iron} & noSR & SupeRVol\\
  \end{tabular}
  }
  \caption{
  Estimated geometry after low resolution training.
  HR and LR denote high and low resolution visualizations of the ground truth geometry, respectively.
  Our framework allows to obtain significantly more detailed reconstructions than IRON\cite{zhang2022iron} even when trained on low resolution images (\nosr).
  The results are further improved in the super-resolution reconstruction (\supervol), which contains some details which are barely visible in the low resolution visualization (LR).}
\label{fig:LR_detailed_geometry}
\end{figure*}

\paragraph{Low resolution training.}
Table~\ref{tab:training} (b) demonstrates quantitatively the effectiveness of correctly modeling the PSF, where we can see that our framework \supervol{} outperforms both IRON~\cite{zhang2022iron} and \nosr{} in terms of geometric quality, as well as novel view synthesis under both colocated and non-colocated lighting.
Regarding novel view synthesis, Figure~\ref{fig:LR_rendering} shows that \supervol{} can recover some crisp details that are barely visible in the individual low resolution images and not properly reconstructed by the competitors.
When it comes to geometric accuracy, as shown in Figure~\ref{fig:LR_detailed_geometry}, the refinement induced from modeling the PSF is not limited to an increased quality of image synthesis, but also has a positive impact on the geometry.
\supervol{} recovers geometric information, which is difficult to see in the low resolution visualization, and at most partially recovered or smoothed by the other approaches, IRON~\cite{zhang2022iron} and \nosr{}.
\begin{figure}[t]
  \centering
  \small
  \newcommand{\mywidthx}{0.145\textwidth}
  \newcolumntype{Y}{ >{\centering\arraybackslash} m{\mywidthx} }
  \newcommand{\tabelt}[1]{\hfil\hbox to 0pt{\hss #1 \hss}\hfil}
  \setlength\tabcolsep{1pt} %
  \begin{tabular}{cYYY}
    \rotatebox[origin=c]{90}{\dogB{}}&
    \includegraphics[width=\mywidthx]{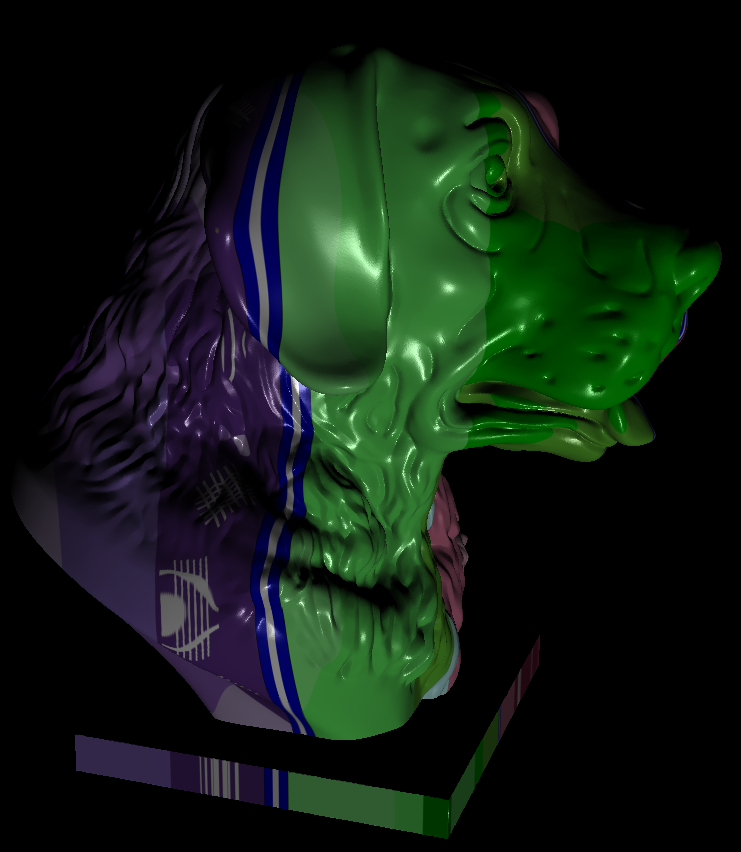}&
    \includegraphics[width=\mywidthx]{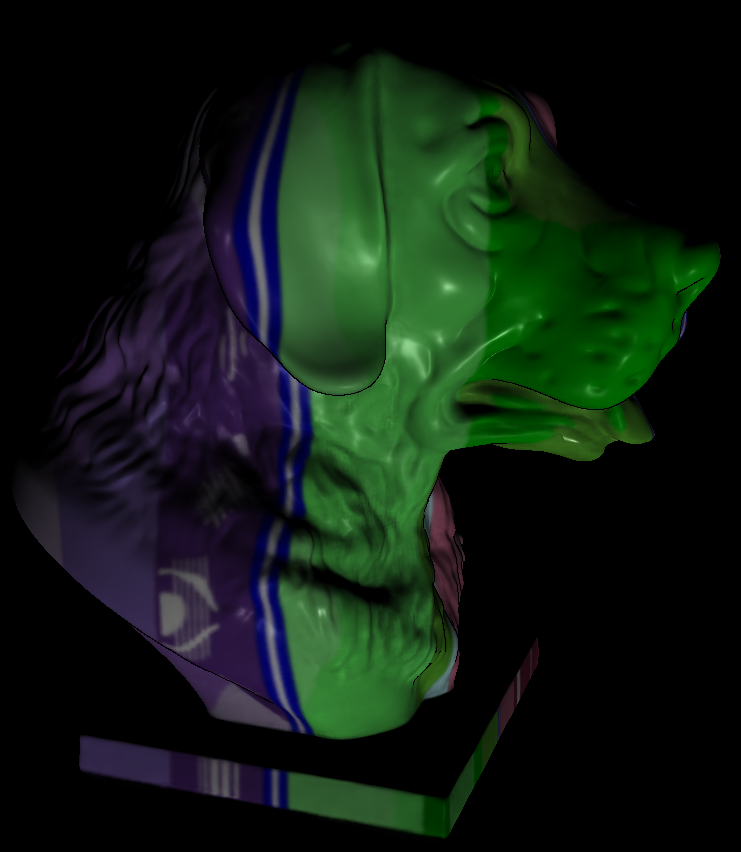}&
    \includegraphics[width=\mywidthx]{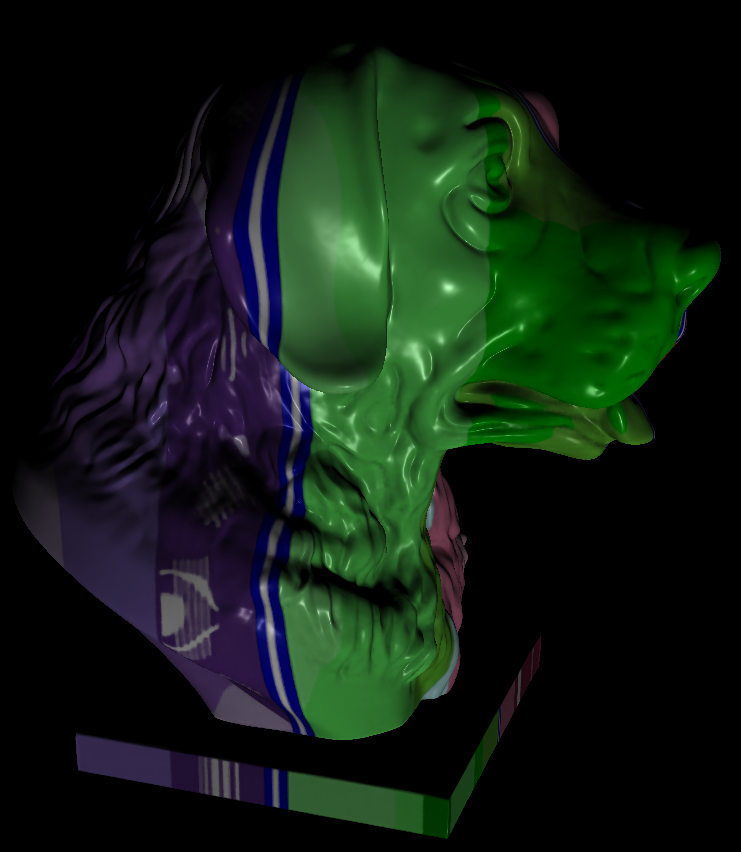}\\
    &Novel lighting & IRON\cite{zhang2022iron} & SupeRVol
  \end{tabular}
  \caption{Generalization to novel non-colocated lighting. 
  Compared to IRON\cite{zhang2022iron}, \supervol{} yields more accurate specularities.
  This demonstrates a better generalization for unseen views and illumination environments.
  }
\label{fig:relighting}
\end{figure}

\begin{figure}[t]
  \centering
  \small
  \newcommand{\mywidthx}{0.145\textwidth}
  \newcolumntype{Y}{ >{\centering\arraybackslash} m{\mywidthx} }
  \newcommand{\tabelt}[1]{\hfil\hbox to 0pt{\hss #1 \hss}\hfil}
  \setlength\tabcolsep{1pt} %
  \begin{tabular}{cYYY}
    \rotatebox[origin=c]{90}{\pony{}}&
    \includegraphics[width=\mywidthx]{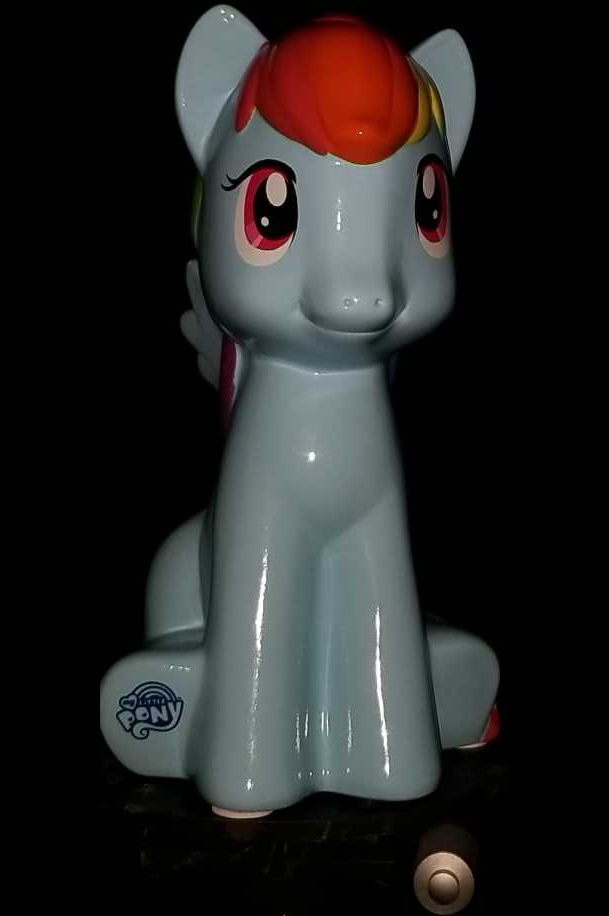}&
    \includegraphics[width=\mywidthx]{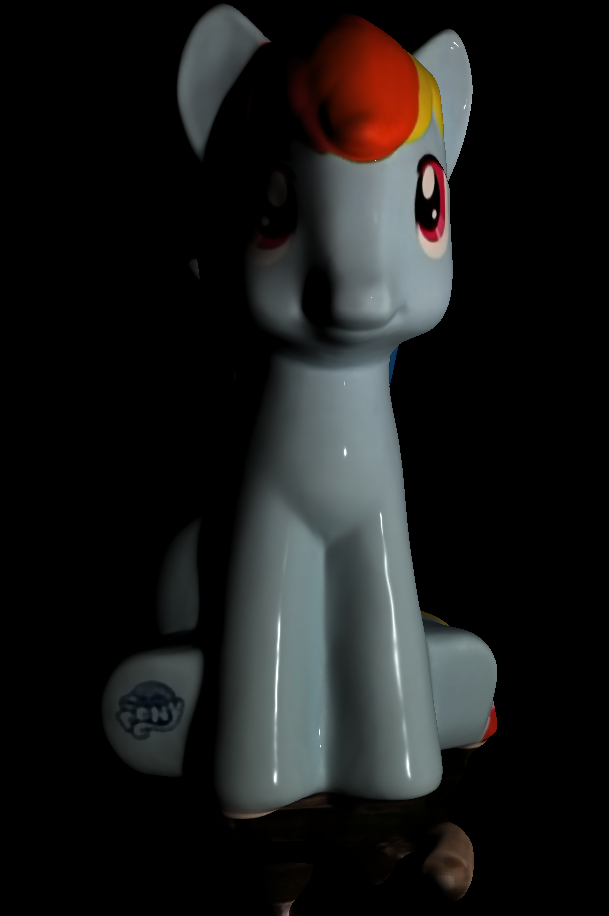}&
    \includegraphics[width=\mywidthx]{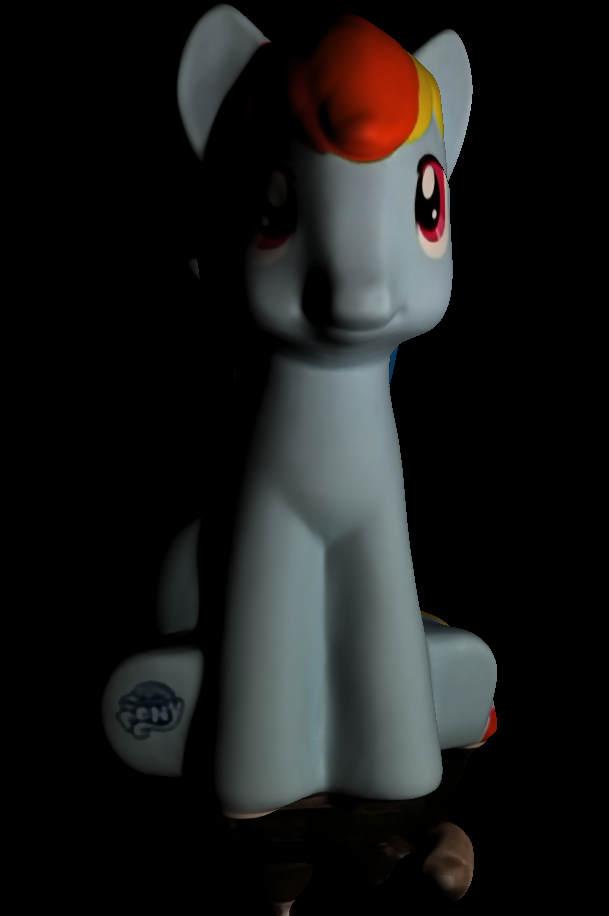}\\
    & Real image & Novel relighting & Lambertian editing\\
  \end{tabular}
  \caption{Novel non-colocated lighting  and material editing of real world data with \supervol{}.
  From the given viewpoint (left), we first moved the light source to the right (middle).
  It can be seen that specularities and shadows moved appropriately.
  Finally, we perform material editing by removing the specular component (right).
  Both together demonstrate the quality of the underlying estimated material.
  A full reconstruction of the real image (left) can be seen in Figure ~\ref{fig:teaser}.
  }
\label{fig:material_editing}
\end{figure}

\paragraph{Generalization to non-colocated relighting.}
Our approach estimates BRDF parameters from a colocated camera-light setup.
This only captures a slice of the BRDF.
Hence, we quantify in Table~\ref{tab:training} under ``synthetic non-colocated'' and visualize in Figure~\ref{fig:relighting} how well our approaches generalizes to non-colocated lighting setups.
As can be seen quantitatively and qualitatively, our proposed model performs better than the state-of-the-art IRON~\cite{zhang2022iron}.
\supervol{} can create realistic specular behavior, while  IRON~\cite{zhang2022iron} shows visible differences to the ground truth.
Finally, Figure~\ref{fig:material_editing} shows both relighting and material editing of \pony{} estimated with SupeRVol.
Although no ground truth is available for comparison, we can clearly see that relighting and editing are intuitively correct.
Hence, it yields a coherent behaviour in terms of specularities and shadows.
This further shows the validity of the estimated material parameters.
For more editing results, see the supplementary material.

\section{Conclusion}
\label{sec:Conclusion}

We propose to enhance suitable neural representations of shape and material parameters with a physical model of the camera that includes the blurring and down-sampling due to the point spread function (PSF).
This leads to a neural approach for recovering shape and material properties at a resolution that is superior to that of baseline methods and even superior to that of the input images.
We carefully motivate the choice of representation including the use of volume rendering over surface rendering and propose %
a generalization of standard approaches via explicitly modeling the PSF.
In qualitative and quantitative studies we demonstrate that the proposed approach outperforms the state-of-the-art and offers a method for high resolution 3D modeling of shape and appearance even from low-resolution input imagery.

\clearpage
{\small
\balance
\bibliographystyle{ieee_fullname}
\bibliography{biblio}
}
\pagebreak
\appendix

\setcounter{section}{0}
\setcounter{equation}{0}
\setcounter{table}{0}

\section{Network Details}

\subsection{Architecture}
As mentioned in the main paper, we use three multilayer perceptrons (MLPs).
One describes the geometry via an SDF, $\sdf_\sdfparam$, one describes the BRDF's diffuse albedo, $\brdfd_{\diffuseparam}$, and one is used for the specular parameters of the material, $\alpha_{\specparam}$.
The MLP of $\sdf_\sdfparam$ consists of $5$ layers of width $512$, with a skip connection at the $4$-th layer.
The MLPs of $\brdfd_{\diffuseparam}$ and $\alpha_{\specparam}$ consist of $4$ layers of width $512$, and $3$ layers of width $256$, respectively.\\
In order to compensate the spectral bias of MLPs \cite{mildenhall2021nerf}, the input is encoded by positional encoding using $6$ frequencies for both $\sdf_\sdfparam$ and $\alpha_{\specparam}$, and $12$ frequencies for $\brdfd_{\diffuseparam}$. 

\subsection{Parameters and Cost Function}
Similarly to \cite{zhang2022iron,yariv2020multiview}, we assume that the scene of interest lies within the unit sphere, which can be achieved by normalizing the camera positions appropriately.
To approximate the Volume rendering integral~\eqref{eq:volume_integral} using~\eqref{eq:discrete_volume_integral}, we use $m=98$ samples which are also used to approximate ~\eqref{eq:weights}, all with the sampling strategy of~\cite{yariv2021volume}.\\
In the following, we distinguish between the ablation study noSR of the main paper and SupeRVol.\\
For SupeRVol, we set the objective's function trade-off parameters $\lambda_1 = \lambda_2=0.1$. Furthermore, in order to approximate the convolution with a Gaussian PSF ~\eqref{eq:psf_image_model}, we use $N_s=25$ in~\eqref{eq:monte_carlo_psf}, and the terms of the objective function~\eqref{eq:data_term} and ~\eqref{eq:eikonal_term} consist of a batch size of $100$ (inside the silhouette) and $1000$, respectively.
For the mask term~\eqref{eq:mask_term} of the objective function, we use the same batch as~\eqref{eq:data_term}, and add around $500$ additional rays outside the silhouette whose rays still intersect with the unit sphere.\\
Concerning the noSR parameters, we set the objective's function trade-off parameters $\lambda_1 = 0.1$, $\lambda_2=0$, i.e. we \textit{turn off} mask supervision, and the terms of the objective function~\eqref{eq:data_term} and~\eqref{eq:eikonal_term} consist of a batch size of $2000$ and $1000$, respectively.\\
Note, that we always normalize each objective function's summand with its corresponding batch size.

\subsection{Training}
We train our networks using the Adam optimizer\cite{kingma2014adam} with a learning rate initialized with $5\e-4$ and decayed exponentially during training to $5\e-5$, except for the MLP $\alpha_{\specparam}$ whose learning rate is constantly equal to $1\e-5$. The remaining parameters are kept to Pytorch's default.\\
We train for $2000$ epochs, which lasts about $2$ days for noSR, and less than $3$ days for SupeRVol using a single NVIDIA P6000 GPU with $24$GB memory and $60$ input images.
For SupeRVol, %
we fix the geometry after the end of the training, and refine the BRDF's parameters using a larger batch size of $700$ -- all within the object's silhouette.

\section{Data Acquisition}
In this section we describe how we generated the datasets used in this paper
\subsection{Synthetic Data}
The synthetic datasets \dogA, \dogB, \girlA, \girlB{} were generated using Blender \cite{blender} and Matlab \cite{MATLAB:2020}, where Blender \cite{blender} is used to render depth, normal and BRDF parameter maps for each viewpoint, and Matlab \cite{MATLAB:2020} is used to render images %
using equation (6) and (7) of the main paper. \\
The low-resolution images, of size $320 \times 240$, are obtained by blurring and downsampling high-resolution images, of size $1280 \times 960$, by a factor four (in each direction).
\subsection{Real World Data}
The real world data of \pony{} and \dragon{} were shared by the authors of~\cite{bi2020deep}, and the real world data of \bird{} and \squirrel{} were created by ourselves.
We use a Samsung Galaxy Note 8 and the application "CameraProfessional"\footnote{\url{https://play.google.com/store/apps/details?id=com.azheng.camera.professional}, accessed 13-th March. 2023, 6.00PM} to generate RAW images as well as the smartphone's images in parallel.
We use the RAW images for our algorithm, and we pre-processed those using Matlab \cite{MATLAB:2020} by following \cite{sumner2014processing}.
Low-resolution images are obtained similarly to synthetic data, which are of size $270 \times 480$ for \pony{} and \dragon{}, and $504 \times 378$ for \bird{} and \squirrel{}.

\section{Novel Renderings}
\begin{figure*}[t]
  \centering
  \small
  \newcommand{\mywidthx}{0.39\textwidth}
  \newcommand{\mywidthc}{0.02\textwidth}
  \newcolumntype{X}{ >{\centering\arraybackslash} m{\mywidthx} }
  \newcolumntype{C}{ >{\centering\arraybackslash} m{\mywidthc} }
  \newcommand{\tabelt}[1]{\hfil\hbox to 0pt{\hss #1 \hss}\hfil}
  \setlength\tabcolsep{1pt} %
  \begin{tabular}{CXX}
    \rotatebox{90}{\pony{}}&
    \includegraphics[width=\mywidthx]{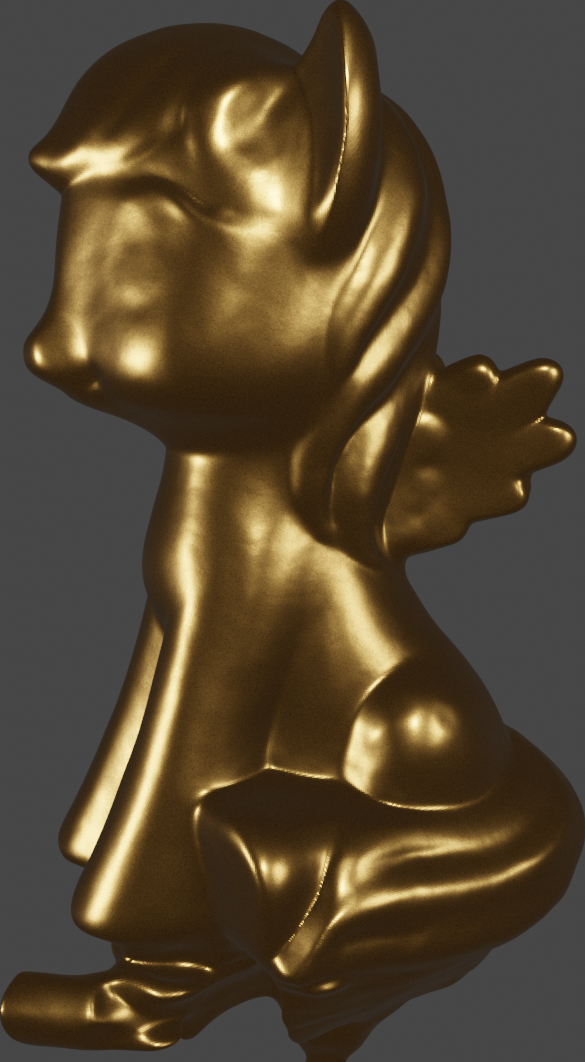}& \includegraphics[width=\mywidthx]{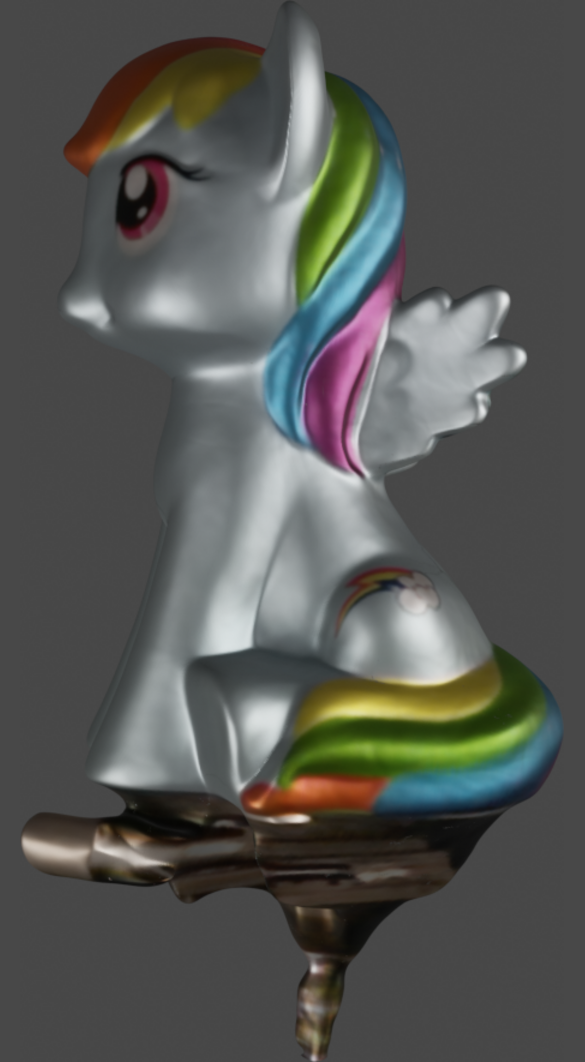}
    \\
    \rotatebox{90}{\bird{}}&
    \includegraphics[width=\mywidthx]{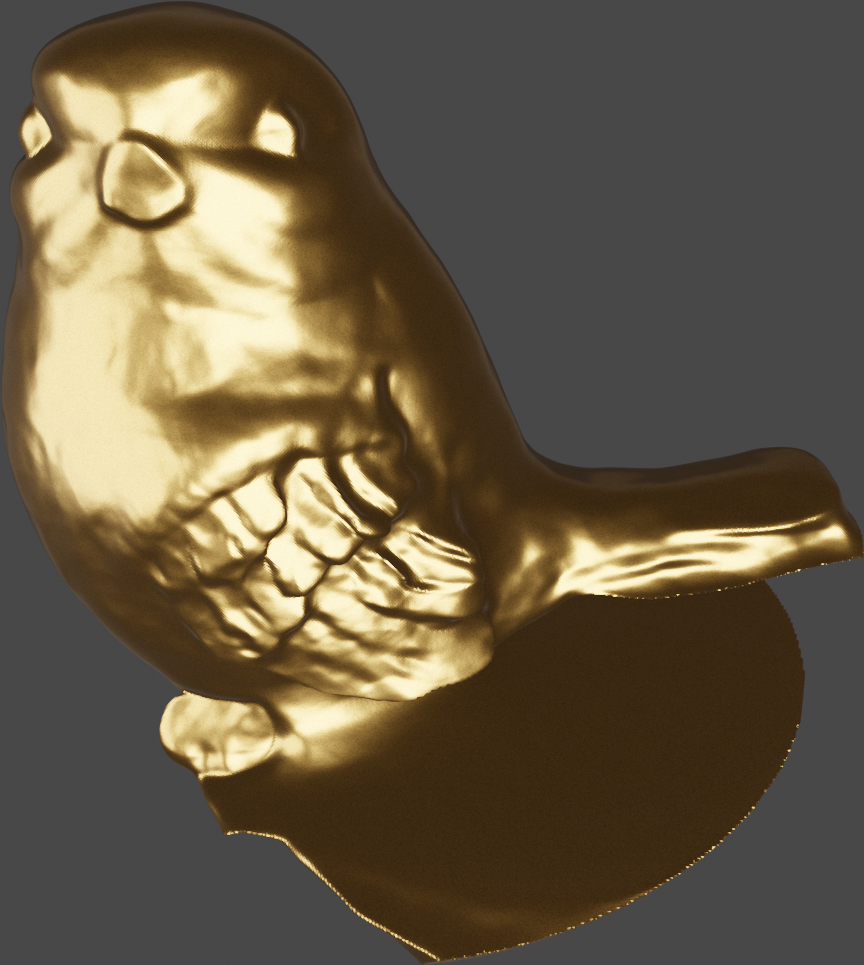}&
    \includegraphics[width=\mywidthx]{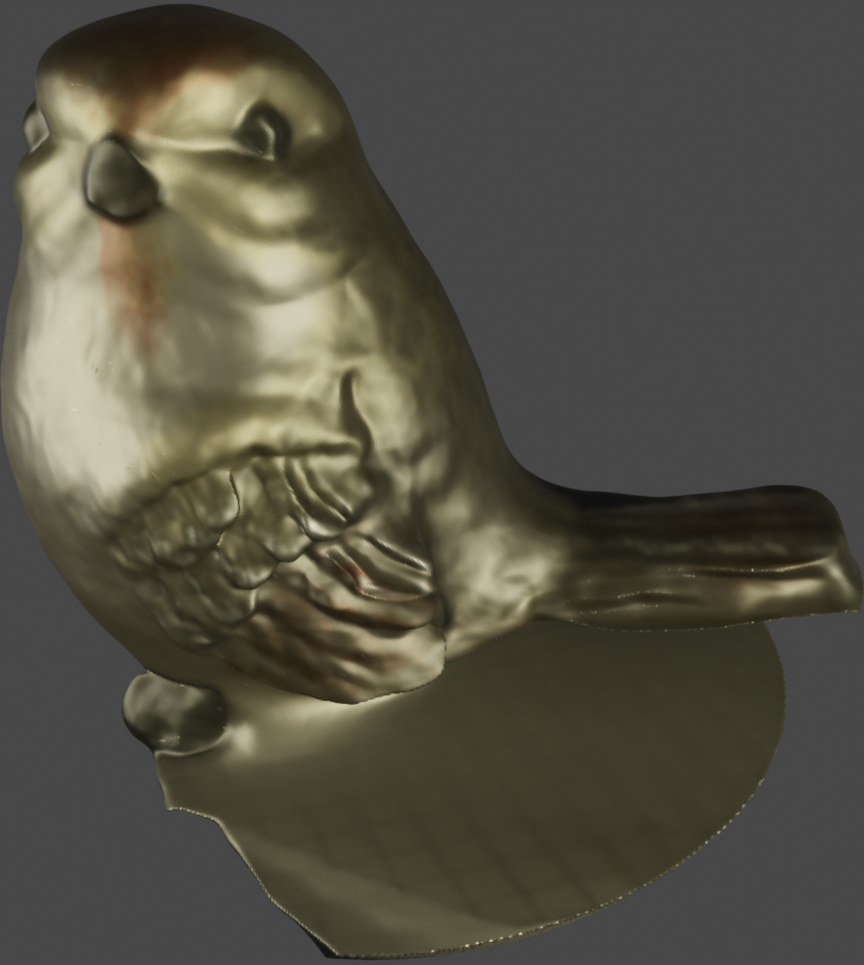}
  \end{tabular}
\caption{Novel rendering of \pony{} and \bird{} dataset. Both shapes where extracted from the learned sdf $\sdf$ using~\cite{10.1145/37402.37422} and their BRDF was altered in Blender\cite{blender}. (left) shows a BRDF simulating gold, (right) uses the estimated diffuse albedo, with a more metallic, rougher and emissive material.}
\label{fig:material_novel}
\end{figure*}

To validate that our approach results in the scene's parameters which can be used to alter the material and visualize it under novel illumination with standard software (Blender \cite{blender}), we show novel renderings in Fig.~\ref{fig:material_novel}.

\end{document}